\documentclass[10pt,twocolumn,letterpaper]{article}

\usepackage{cvpr}
\usepackage{times}
\usepackage{epsfig}
\usepackage{graphicx}
\usepackage{amsmath}
\usepackage{amssymb}

\usepackage{times}
\usepackage{epsfig}
\usepackage{graphicx}
\usepackage{float}
\usepackage{wrapfig}
\usepackage{amsmath,amssymb,amsthm}
\usepackage{algorithm,algorithmicx,algpseudocode}
\usepackage{bm,xspace}
\usepackage{comment}
\usepackage{verbatim}
\usepackage{multirow}
\usepackage{balance}
\usepackage{url}
\usepackage{booktabs}
\usepackage{etoolbox,siunitx}
\usepackage{calc}
\usepackage{pifont,hologo}
\usepackage[usenames, dvipsnames]{color}
\usepackage{nicefrac}

\newcommand{\ra}[1]{\renewcommand{\arraystretch}{#1}}

\usepackage[super]{nth}
\usepackage{nicefrac}
\robustify\bfseries

\def\CC{\mathbf{C}}

\def\XX{\mathbf{X}}
\def\YY{\mathbf{Y}}

\DeclareMathSymbol{@}{\mathord}{letters}{"3B}

\newcommand\mypara[1]{\vspace{1mm}\noindent\textbf{#1}}

\def\latex/{\LaTeX}
\def\bibtex/{\hologo{BibTeX}}

\usepackage[pagebackref=true,breaklinks=true,letterpaper=true,colorlinks,bookmarks=false]{hyperref}

\cvprfinalcopy 

\def\cvprPaperID{1399} 

\ifcvprfinal\fi
\begin{document}

\title{Fully Automatic Video Colorization with Self-Regularization and Diversity}

\author{Chenyang Lei\\
HKUST\\
\and
Qifeng Chen\\
HKUST\\
}

\maketitle

\begin{abstract}
We present a fully automatic approach to video colorization with self-regularization and diversity. Our model contains a colorization network for video frame colorization and a refinement network for spatiotemporal color refinement. Without any labeled data, both networks can be trained with self-regularized losses defined in bilateral and temporal space. The bilateral loss enforces color consistency between neighboring pixels in a bilateral space and the temporal loss imposes constraints between corresponding pixels in two nearby frames. While video colorization is a multi-modal problem, our method uses a perceptual loss with diversity to differentiate various modes in the solution space. Perceptual experiments demonstrate that our approach outperforms state-of-the-art approaches on fully automatic video colorization.
\end{abstract}

%


\begin{figure*}[t!]
\centering
\begin{tabular}{@{}c@{\hspace{2mm}}c@{\hspace{1mm}}c@{\hspace{2mm}}c@{\hspace{1mm}}c@{\hspace{2mm}}c@{}}
\rotatebox{90}{\small \hspace{10mm} Input}
&\includegraphics[width=0.23\linewidth]{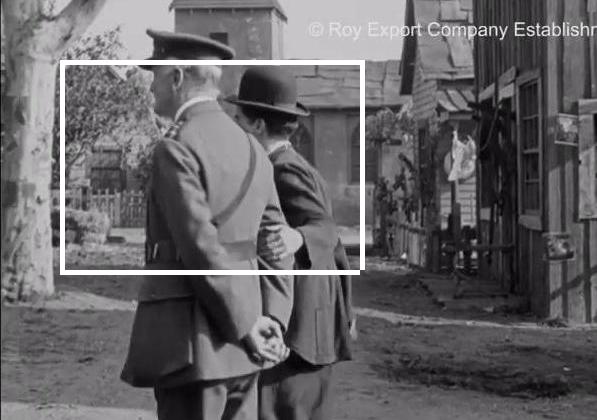}&
\includegraphics[width=0.23\linewidth]{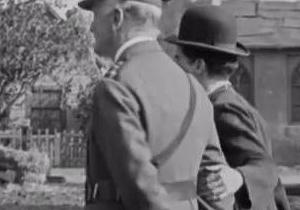}&
\includegraphics[width=0.23\linewidth]{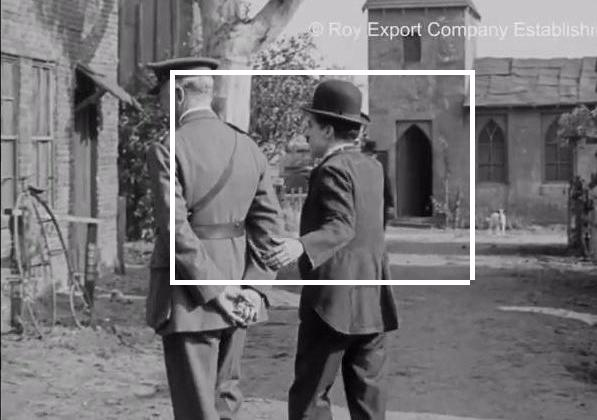}&
\includegraphics[width=0.23\linewidth]{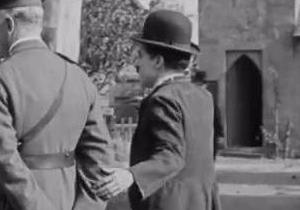}\\
\rotatebox{90}{\small \hspace{3mm} Zhang et al.~\cite{Zhang2016}}
&\includegraphics[width=0.23\linewidth]{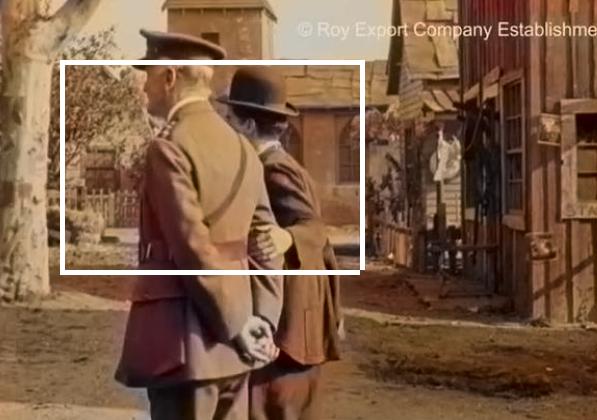}&
\includegraphics[width=0.23\linewidth]{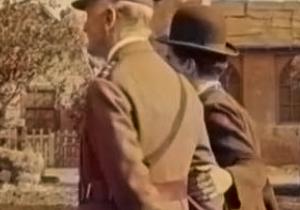}&
\includegraphics[width=0.23\linewidth]{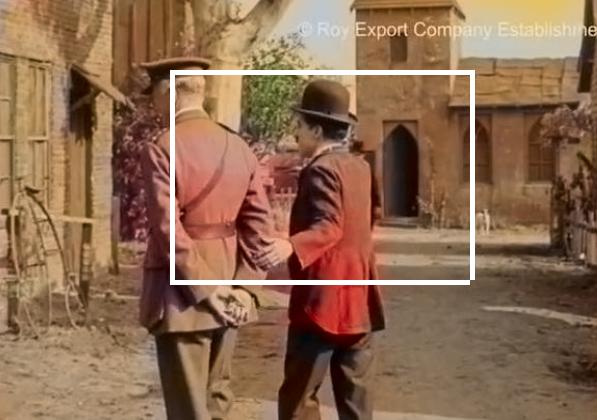}&
\includegraphics[width=0.23\linewidth]{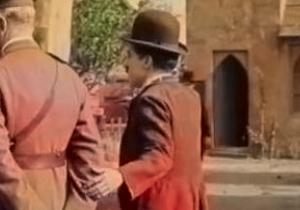}\\
\rotatebox{90}{\small \hspace{3mm}Iizuka et al.~\cite{Iizuka2016}}
&\includegraphics[width=0.23\linewidth]{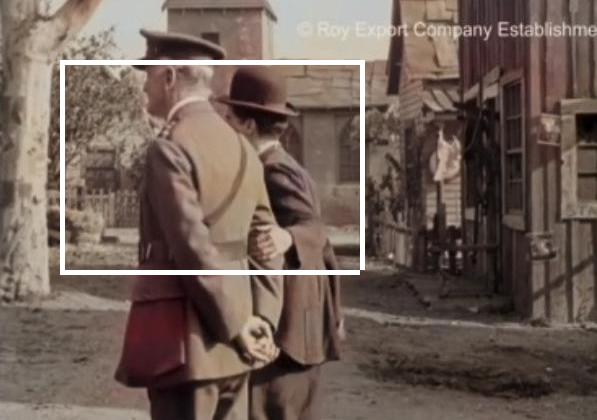}&
\includegraphics[width=0.23\linewidth]{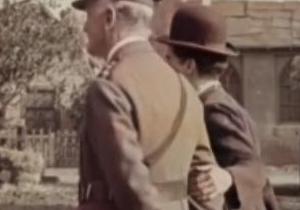}&
\includegraphics[width=0.23\linewidth]{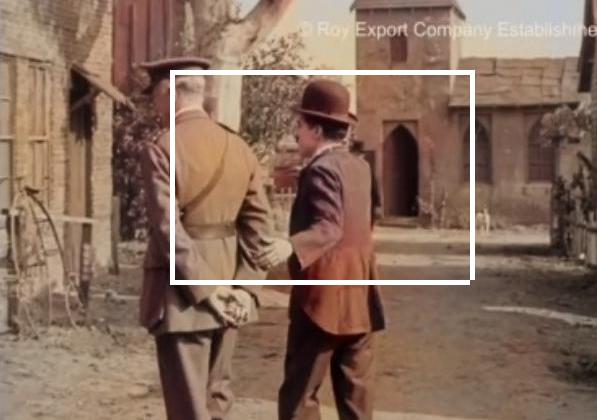}&
\includegraphics[width=0.23\linewidth]{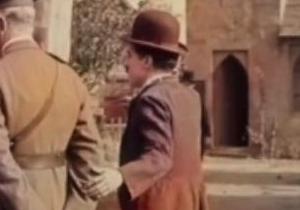}\\
\rotatebox{90}{\small \hspace{11mm} Ours}
&\includegraphics[width=0.23\linewidth]{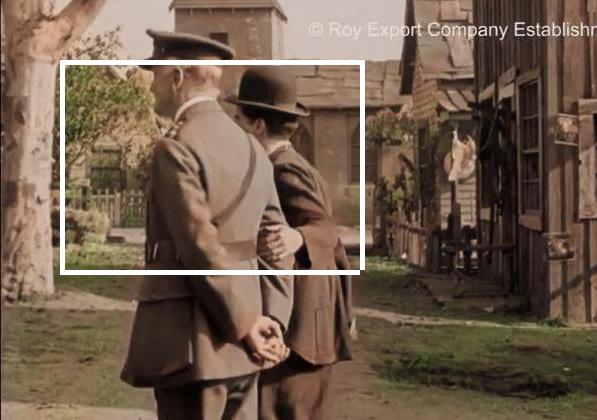}&
\includegraphics[width=0.23\linewidth]{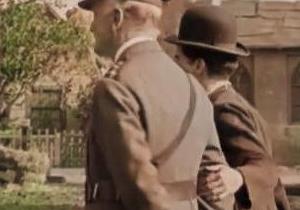}&
\includegraphics[width=0.23\linewidth]{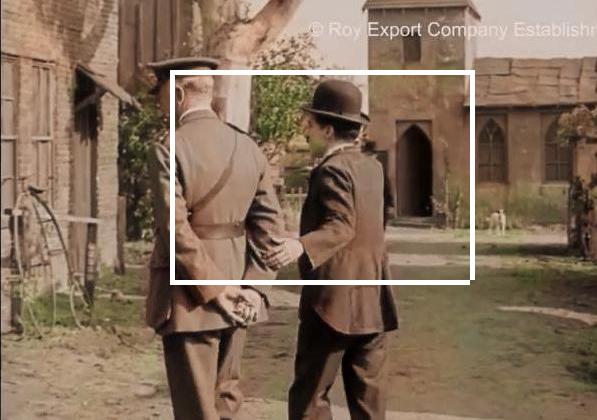}&
\includegraphics[width=0.23\linewidth]{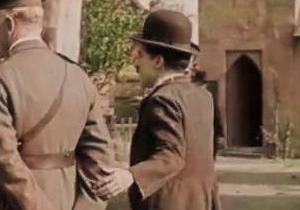}\\
&\multicolumn{2}{c}{Frame 1} & \multicolumn{2}{c}{Frame 2} \\
\end{tabular}
\vspace{1mm}
\caption{Two colorized video frames by Zhang et al. \cite{Zhang2016}, Iizuka et al. \cite{Iizuka2016}, and our approach on the classic film \textit{Behind the Screen} in 1916 by Charlie Chaplin. State-of-the-art image colorization methods may not perform well on video colorization. The temporal inconsistency between the colorized video frames by Zhang et al. \cite{Zhang2016} and Iizuka et al. \cite{Iizuka2016} is obvious. More results of classic film colorization are shown in the supplement.}
\label{fig:teaser}
\end{figure*}

\section{Introduction}
There exist numerous classic films and videos in black-and-white. It is desirable for people to watch a colorful movie rather than a grayscale one. \textit{Gone with the Wind} in 1939 is one of the first colorized films and is also the all-time highest-grossing film adjusted for inflation \cite{boxoffice2014}. Image and video colorization can also assist other computer vision applications such as visual understanding \cite{Larsson2017} and object tracking \cite{Vondrick2018}.

Video colorization is highly challenging due to its multi-modality in the solution space and the requirement of global spatiotemporal consistency. First, it is not reasonable to recover the ground-truth color in various cases. For example, given a grayscale image of a balloon, we can not predict the correct color of the balloon because it may be yellow, blue, and so on. Instead of recovering the underlying color, we aim to generate a set of colorized results that look natural. Second, it often does not matter what color we assign to a region (i.e. a balloon), but the whole region should be spatially consistent. Third, video colorization is also inherently more challenging than single image colorization since temporal coherence should be also enforced. Image colorization methods usually do not generalize to video colorization. In Figure~\ref{fig:teaser}, we show some results of our approach and two state-of-the-art image colorization methods on classic film colorization. 

Colorization of black-and-white images has been well studied in the literature \cite{Levin2004,Cheng2015,Zhang2016,Larsson2016}. Colorization methods in the early days are mostly user-guided approaches that solve an objective function to propagate user input color scribbles to other regions \cite{Levin2004,Qu2006}. These approaches require users to provide sufficient scribbles on the grayscale image. On the other hand, researchers explore automatic image colorization with deep learning models. 
Some deep learning based approach for image colorization defines a classification based loss function with hundreds of discrete sampled points in chrominance space \cite{Zhang2016,Larsson2016}. However, the colorized image often exhibits evident discretization artifacts. To tackle this challenge, we suggest using a perceptual loss function combined with diversity.  Our approach does not rely on sampling a discrete set of color in chrominance space and thus avoids discretization artifacts in the colorized video.

We may apply image colorization methods to colorize video frames independently, but the overall colorized video tends to be temporally inconsistent. Recently, Lai et al. \cite{Lai2018} proposed a framework to enhance temporal coherence of a synthesized video where each frame is processed independently by an image processing algorithm such as colorization. However, this is a post-processing step and its performance is dependant on an image colorization approach that does not utilize multiple-frame information. Propagation-based video colorization methods require some colorized frames as reference to propagate the color of the given reference frames to the whole video \cite{Meyer2018,Vondrick2018}, but colorizing some frames also requires non-trivial human effort. Also, the quality of the colorized video frames decays quickly when the future frames are different from the reference frames. In this paper, we study the problem of automatic video colorization without both labeled data and user guidance.

We propose a self-regularized approach to automatic video colorization with diversity. We regularize our model with nearest neighbors in both bilateral and temporal spaces, and train the model with a diversity loss to differentiate different modes in the solution space. The self-regularization encourages information propagation between pixels expected to have similar color. Specifically, we can build a graph with explicit pairwise connections between pixels by finding $K$ nearest neighbors in some feature space or following the optical flow. By enforcing pairwise similarity between pixel pairs, we can preserve spatiotemporal color consistency in a video. Our model is also capable of generating multiple diverse colorized videos with a diversity loss \cite{Li2018}. We further suggest a simple strategy to select the most colorful video among all colorized videos.


We conduct experiments to compare our model with state-of-the-art image and video colorization approaches. The results demonstrate that our model can synthesize more natural colorized videos than other approaches do. We evaluate the performance on PSNR and LPIPS \cite{Zhang2018}, and conduct perceptual comparison by a user study. Furthermore, controlled experiments show that our self-regularization and diversity are critical components in our model.

\begin{figure*}[t!]
\centering
\includegraphics[width=\linewidth]{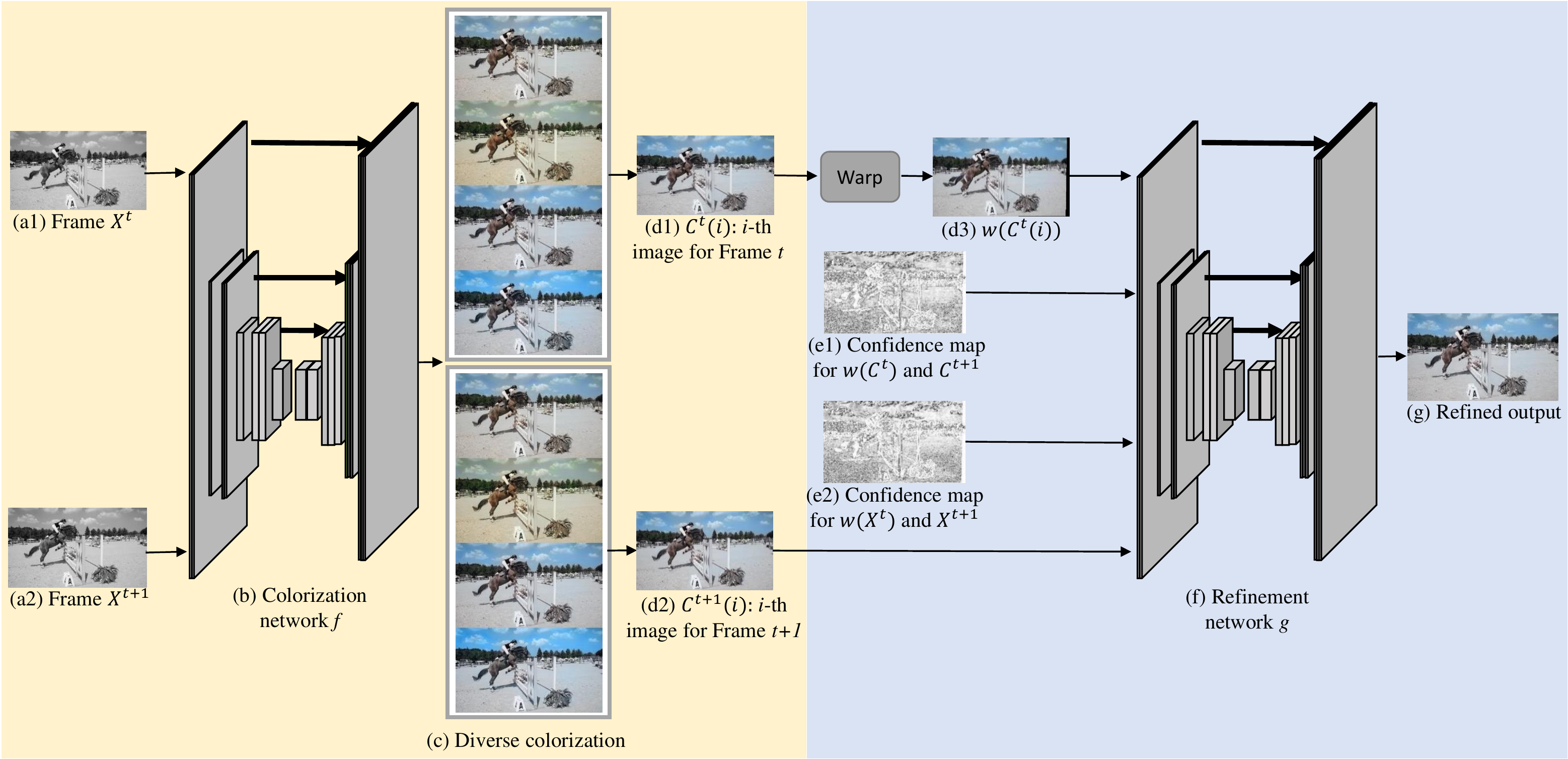}
\vspace{1mm}
\caption{The overall architecture of our model. The colorization network $f$ is designed to colorize each grayscale video frame, and produces multiple colorization candidate images. Taking $i$-th colorized candidate images from Frame $t$ and Frame $t+1$ as well as two confidence maps, the refinement network $g$ will output a refined video frame for Frame $t$.}
\label{fig:overview}
\end{figure*}

\section{Related Work}
In this section, we briefly review the related work in image and video colorization.

\mypara{User-guided Image Colorization.}
The most classical approaches on image colorization are based on optimization that requires user input on part of the image to propagate the provided colors on certain regions to the whole image \cite{Levin2004,Qu2006,Luan2007,Chen2012,Yatziv2006}. Levin et al. \cite{Levin2004} propose optimization based interactive image colorization by solving a quadratic cost function under the assumption that similar pixels in space-time should have similar colors. Zhang et al. \cite{Zhang2017} present a deep learning based model for interactive image colorization.

Instead of requiring user scribbles, exemplar-based colorization approaches take a reference image as additional input \cite{Welsh2002,Ironi2005,Liu2008,Charpiat2008,Chia2011,Gupta2012}. The reference image should be semantically similar to the input grayscale image to transfer the color from the reference image to the input image.  A recent approach by He et al. \cite{He2018} combines deep learning and exemplars in image colorization and achieves the state-of-the-art performance. In this work, we are interested in  fully automatic colorization approach that requires neither user input nor reference images.



\mypara{Automatic Image Colorization.}
The most prominent work on fully automatic image colorization is deep learning based approaches that do not require any user guidance \cite{Cheng2015,Iizuka2016,Zhang2016,Larsson2016,Deshpande2017}. Cheng et al. \cite{Cheng2015} propose the first deep neural network model for fully automatic image colorization. Some deep learning approaches use a classification network that classifies each pixel into a set of hundreds of chrominance samples in a LAB or HSV color space to tackle to the multi-modal nature of the colorization problem \cite{Zhang2016,Larsson2016}. However, it is difficult to sample densely in the two-dimensional chrominance with hundreds of points. Thus we propose to use a perceptual loss with diversity \cite{Li2018} to avoid the discretization problem.


\mypara{Video Colorization.}
Most contemporaneous work on video colorization is designed to propagate the color information from a color reference frame or sparse user scribbles to the whole video \cite{Yatziv2006,Vondrick2018,Meyer2018,SwitchableTN, jampani:vpn:2017}. On the other hand, Lai et al. \cite{Lai2018} propose an approach to enforce stronger temporal consistency of a video generated frame by frame by an image processing algorithm such as colorization. To the best of our knowledge, there are no deep learning models dedicated to fully automatic video colorization. We can definitely apply an image colorization method to colorize each frame in a video, but the resulted video is usually temporally incoherent. In this paper, we present a dedicated deep learning model for automatic video colorization that encourages spatiotemporal context propagation and is capable of generating a set of different colorized videos.

\section{Overview}
Consider a sequence of grayscale video frames $\XX=\left\{X^1,\hdots,X^n\right\}$. Our objective is to train a model that automatically colorizes $\XX$ such that the colorized video is realistic. In our framework, neither user guidance neither color reference frames are needed. Before we describe our approach, we characterize two desirable properties of our fully automatic video colorization approach.
\begin{itemize}
    \item \textbf{Spatiotemporal color consistency.} Within a video frame, multiple pixels can share a similar color. For example, all the pixels on a wall should have the same color, and all the grass should be green. Establishing nonlocal pixel neighbors (i.e. two pixels on the same wall) for color consistency can improve the global color consistency of a colorized video. Note that colorizing video frames independently can result in a temporally inconsistent video, and thus we can establish temporal neighbors between two frames to enforce temporal coherence.
    \item \textbf{Diverse colorization.} Most existing work on image or video colorization only generates one colorization result.  It is desirable for our model to output a set of diverse set of colorized videos, as colorization is a one-to-many problem. In our model, we use a perceptual loss with diversity to differentiate different modes in the solution space.
\end{itemize}

Figure \ref{fig:overview} illustrates the overall structure of our model. Our proposed framework contains two networks that are trained to work in synergy. The first one is the colorization network $f(X^t;\theta_f)$ that outputs a colorized video frame given a grayscale video frame $X^t$. The network $f$ is self-regularized with color similarity constraints defined on $K$ nearest neighbors in the bilateral space $(r,g,b,\lambda x,\lambda y)$ where $(r,g,b)$ represents the pixel color, $(x,y)$ indicates the pixel location, and $\lambda$ is a weight that balances the pixel color and location. We use $K=5$ in our experiments. The second one is the refinement network $g(C^s, C^t;\theta_g)$ designed to refine the current colorized video $\CC$ by enforcing stronger temporal consistency. The network $g$ propagates information between two nearby frames $C^s$ and $C^t$. At the test time, $g$ can be applied multiple times to the colorized video to achieve long-term consistency.

Furthermore, our approach can produce a diverse set of colorized videos, regularized by the diversity loss introduced by Li et al. \cite{Li2018}. We find that our diversity loss also stabilizes the temporal consistency of the colorized video. Combining the self-regularization and the diversity loss, we obtain the overall loss function to train our model:
\begin{equation}
L_{self} + L_{diversity},
\end{equation}
where $L_{self}$ represents the loss to regularize color similarity between pixel neighbors in a bilateral space and a temporal domain, and $L_{diversity}$ is a perceptual loss function with diversity.

\section{Self-Regularization}
\subsection{Self-regularization for colorization network}
Consider colorizing a textureless balloon. Although it is nearly impossible to infer the underlying color of the balloon from a grayscale video frame, we somehow believe that all the pixels on the balloon are similar. We can find out pixel pairs expected to be similar, and enforce color similarity on these pixel pairs when training our model.

To establish pixel pairs with similar color in a video frame, we perform the $K$ nearest neighbor (KNN) search in a bilateral space $(r,g,b,\lambda x,\lambda y)$ on the ground-truth frame during training. We expect that two pixels with similar color and spatial locations imply that our colorized video should also have a similar color for these two pixels. A similar KNN strategy is also presented in KNN matting \cite{Chen2013}. Suppose $\XX=\{X^1,\hdots,X^n\}$ is the input grayscale video and $\YY=\{Y^1,\hdots,Y^n\}$ is the ground-truth color video, our bilateral loss for self-regularization is 
\begin{equation}
    L_{bilateral}(\theta_f) = \sum_{i=1}^n{\sum_{(p,q)\in \mathcal{N}_{Y^t}}{\|f_p(X^t;\theta_f)-f_q(X^t;\theta_f)\|_1}},
\end{equation}
where $\mathcal{N}_{Y^t}$ is the KNN graph build on the ground-truth color frame $Y_i$, and $f_p(X^t;\theta_f)$ indicates the color of pixel $p$ on the colorized video frame $f(X^t;\theta_f)$.

A simple temporal regularization term $L_{temporal}^f(\theta_f)$ can be defined on $f$:
\begin{equation}
\sum_{t=1}^{n-1}{\|\left(f(X^t;\theta_f)-\omega_{t+1\rightarrow t}(f(X^{t+1};\theta_f))\right)\odot M_{t+1\rightarrow t}\|_1},
\end{equation}
where $\omega_{i+1\rightarrow i}$ is an warping operator that warps an image from Frame $t+1$ to Frame $t$ according to the optical flow from  $X^{t+1}$ to $X^t$. Given the optical flow $f_{t+1->t}$ from frame $t+1$ to frame $t$, we use backward warping to obtain a binary mask $M_{t+1->t}$ that indicates non-occluded pixels (invisible in Frame $t+1$).

\begin{figure*}[t!]
\centering
\begin{tabular}{@{}c@{\hspace{1mm}}c@{\hspace{1mm}}c@{\hspace{1mm}}c@{\hspace{1mm}}c@{}}
\rotatebox{90}{\small \hspace{5mm} Output 1}&
\includegraphics[width=0.24\linewidth]{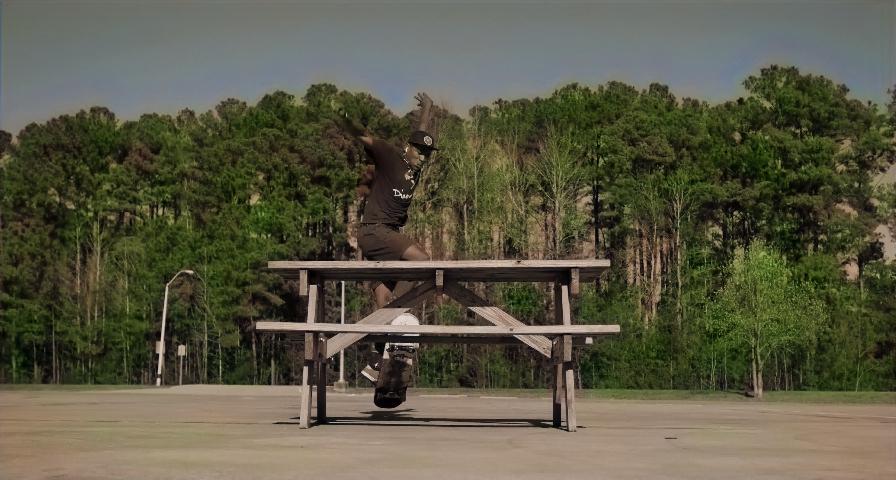}&
\includegraphics[width=0.24\linewidth]{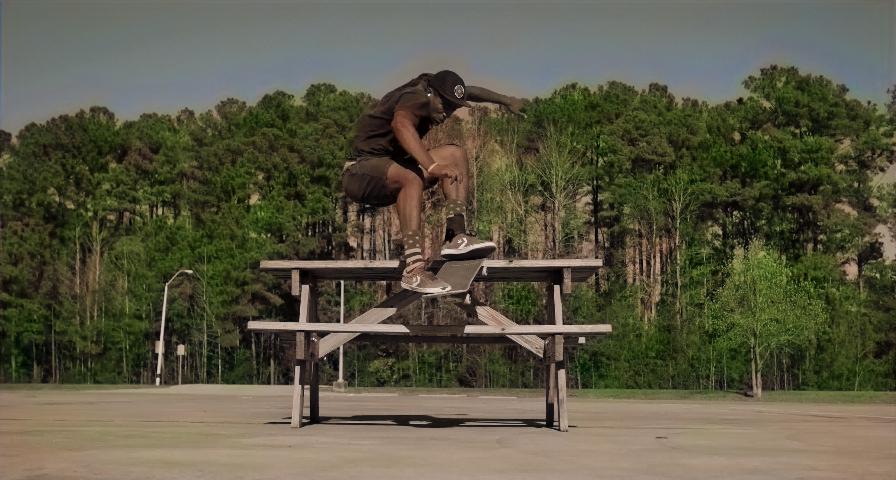}&
\includegraphics[width=0.24\linewidth]{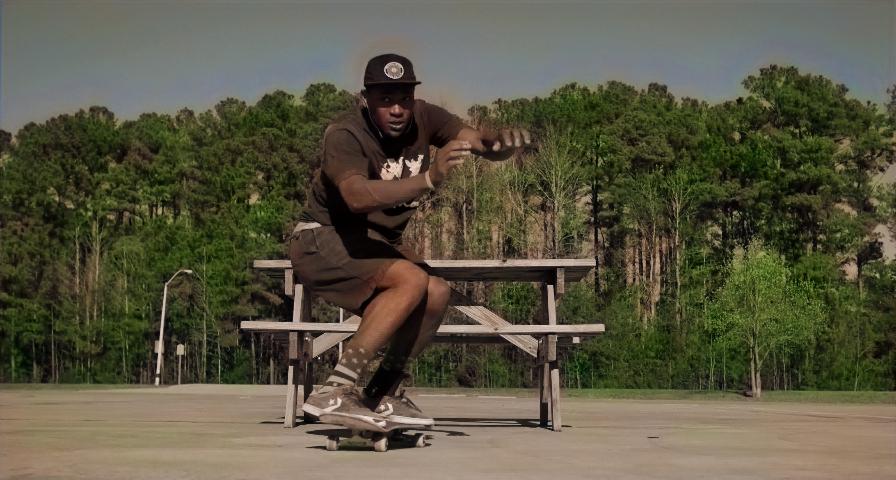}&
\includegraphics[width=0.24\linewidth]{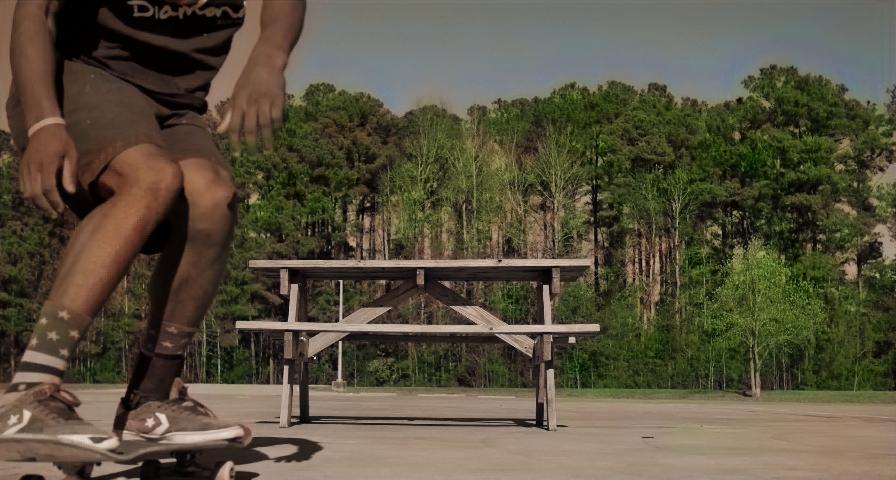}\\
\rotatebox{90}{\small \hspace{5mm} Output 2}& \includegraphics[width=0.24\linewidth]{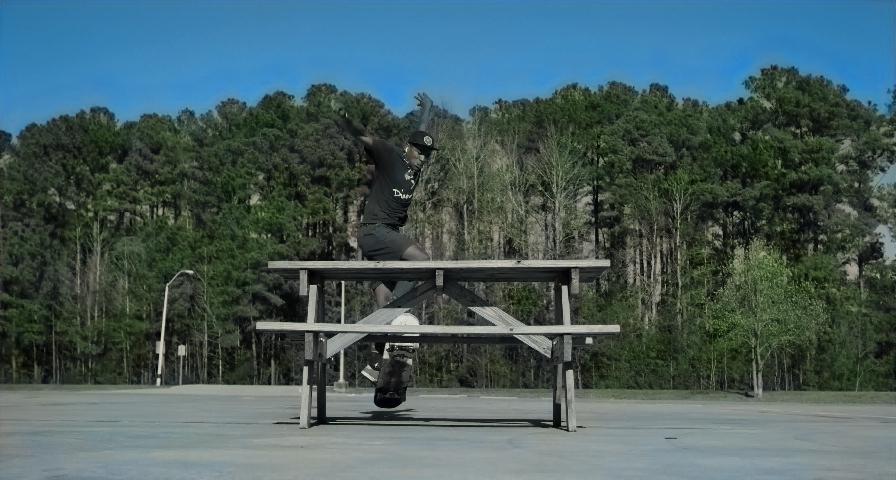}&
\includegraphics[width=0.24\linewidth]{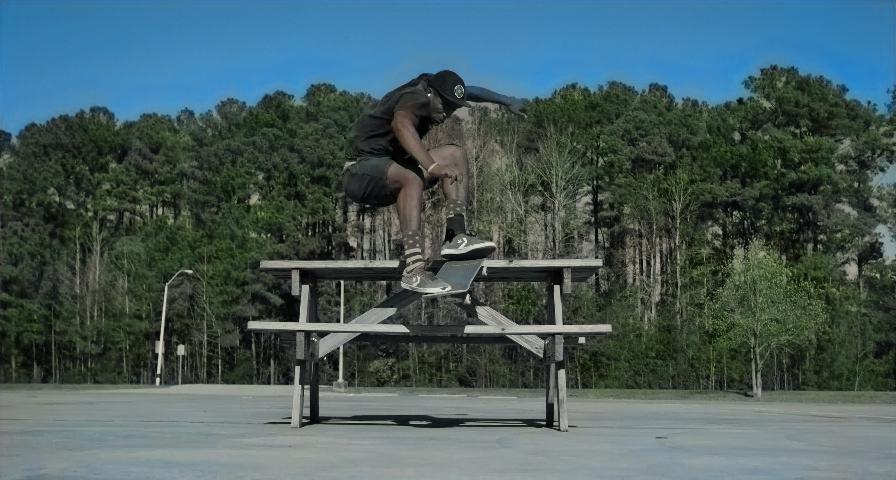}&
\includegraphics[width=0.24\linewidth]{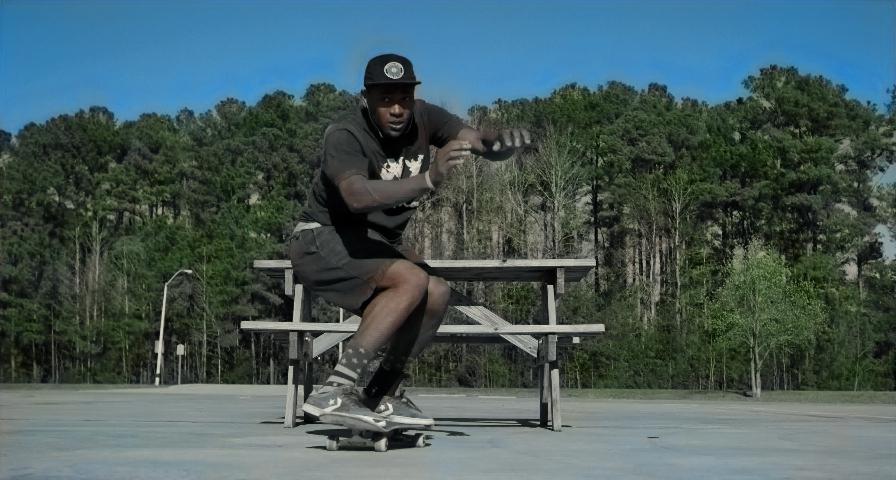}&
\includegraphics[width=0.24\linewidth]{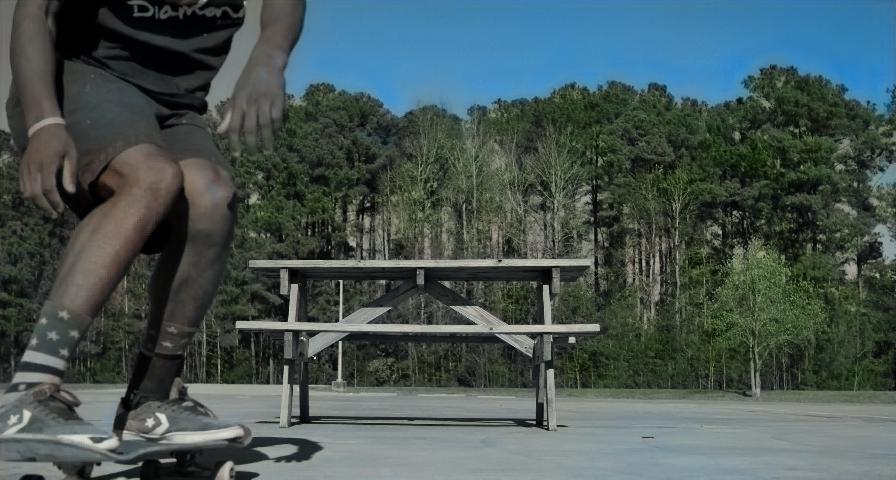}\\
\rotatebox{90}{\small \hspace{5mm} Output 3}&
\includegraphics[width=0.24\linewidth]{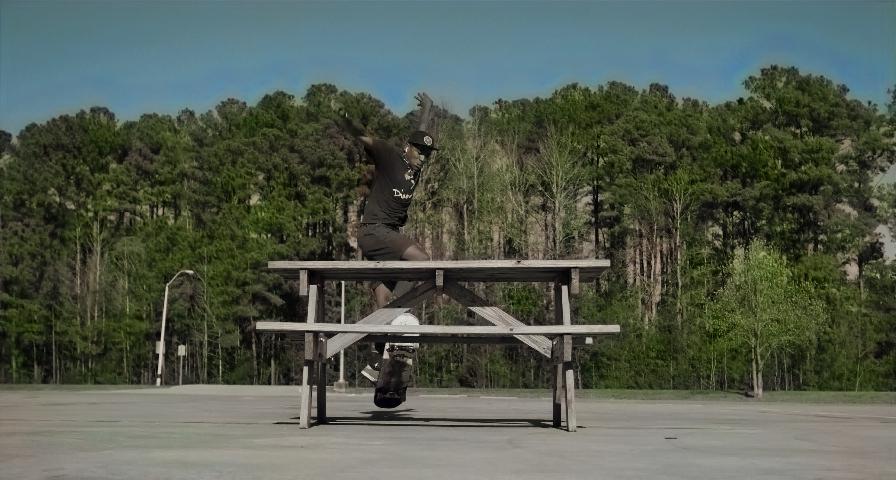}&
\includegraphics[width=0.24\linewidth]{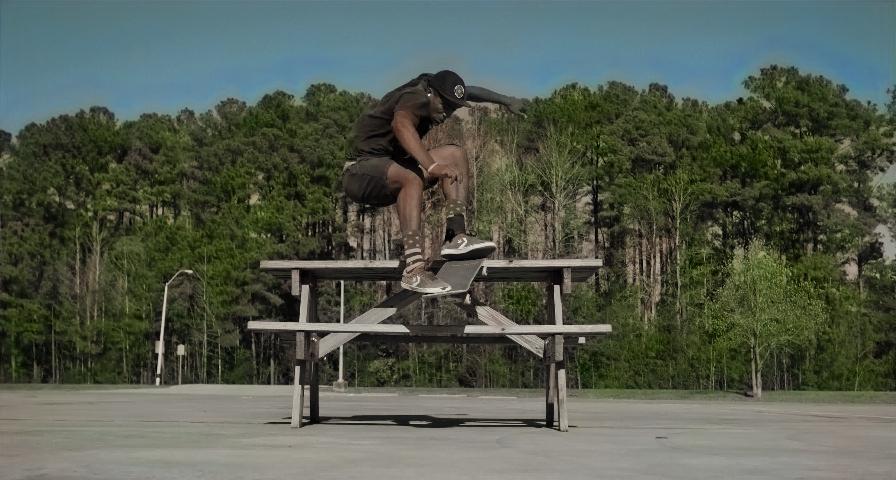}&
\includegraphics[width=0.24\linewidth]{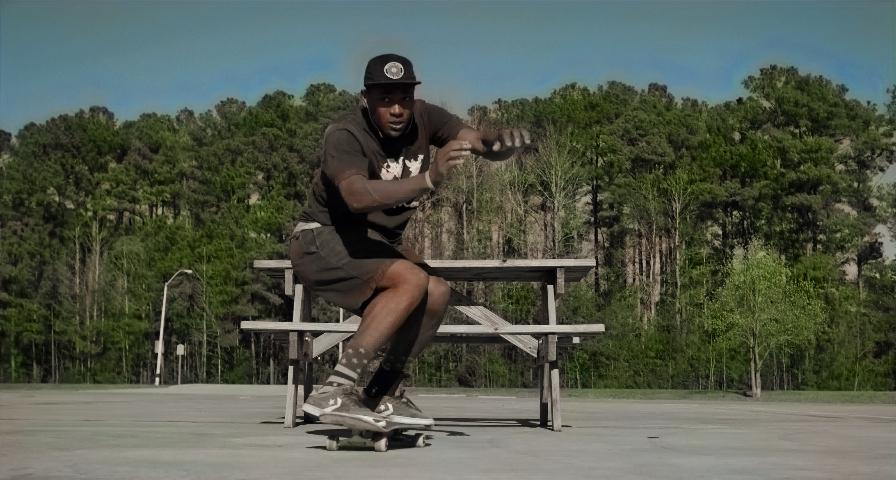}&
\includegraphics[width=0.24\linewidth]{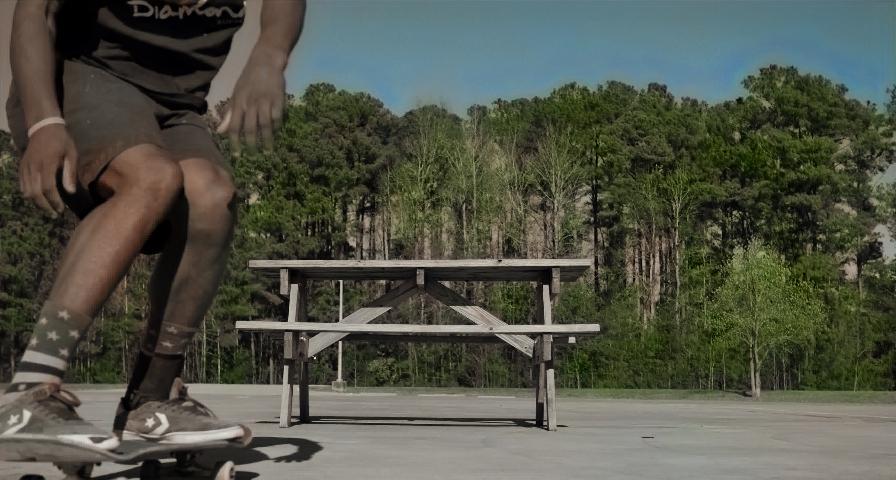}\\

&Frame 1 & Frame 2 & Frame 3 & Frame 4 \\
\end{tabular}
\vspace{1mm}
\caption{Four frames of three different videos colorized by our approach with diversity. Our approach is able to colorize videos in different ways. In general, different videos exhibit different global styles.}
\label{fig:diversity}
\end{figure*}

\subsection{Confidence-based refinement network}
In our model, a confidence-based refinement network $g$ is used to enforce stronger temporal consistency. Temporal inconsistency appears when corresponding pixels in two frames do not share similar colors. We use confidence maps to indicate whether the color of a pixel is inconsistent or inaccurate. Given a current colorized video $\CC=\{C^1,\hdots,C^n\}$, the temporal inconsistency when warping Frame $t$ to Frame $s$ can be translated into a confidence map with weights in the range of $[0,1]$:
\begin{equation}
     W_{t\rightarrow s}(C^t,C^s)=\max( \mathbf{1} - \alpha|C^s - \omega_{t\rightarrow s}(C^t)|\odot M_{t\rightarrow s},\mathbf{0}),
\end{equation}
where $\alpha$ is a hyper-parameter that controls the sensitivity of temporal inconsistency and we use $\alpha=15$.

Thus, for each colorized frame $C^s$, the refinement network $g$ can use another nearby frame $C^t$ along with the computed confidence maps to refine $C^s$. The input to $g$ includes $C^s$, $\omega_{t\rightarrow s}(C^t)$, $W_{t\rightarrow s}(C^t,C^s)$, and $W_{t\rightarrow s}(X^t,X^s)$ that is the confidence map defined on the input grayscale image pairs. $g$ outputs a refined video frame for $C^s$.

\paragraph{Training.} To train the refinement network $g$, we sample two neighboring frames $s$ and $t$ such that $|s-t|\leq\lambda$ where $\lambda$ specifies the window size for temporal refinement. We find $\lambda=1$ is enough in our model. Then we optimize the following temporal regularization loss  for $\theta_g$:

\begin{align}
&L_{temporal}^{g}(\theta_g)=\nonumber\\
&\sum_{1\leq |s-t| \leq \lambda}{\|g(f(X^s;\theta_f),f(X^t;\theta_f);\theta_g)-Y^s\|_1}.
\end{align}

In summary, our self-regularization loss $L_{self}$ is defined as 
\begin{equation}
L_{bilateral}(\theta_f)+L_{temporal}^f(\theta_f)+L_{temporal}^{g}(\theta_g).
\end{equation}

\paragraph{Inference.} During the inference, we can apply $g$ to refine each frame using the left $\lambda$ frames and the right $\lambda$ frames. If we perform this temporal refinement multiple times, we indirectly use the information from non-local frames to refine each frame.

\section{Diverse Colorization}
Video colorization is essentially a one-to-many task as there are multiple feasible colorized videos given the same grayscale input. Generating a diverse set of solutions can be an effective way to tackle this multi-modality challenge. Inspired by the ranked diversity loss proposed by Li et al. \cite{Li2018}, we propose to generate multiple colorized videos to differentiate different solution modes. Besides, the diversity loss also contributes a lot to the temporal coherence because it reduces the ambiguity of colorization by generating several modes. 

Suppose we generate $d$ different solutions in our model. The network $f$ should be modified to generate $d$ images as output. The diversity loss imposed on $f$ is,

\begin{align}
L_{diversity}(\theta_f)=&\sum_{t=1}^n{\min_i\{\|\phi(C^t(i))-\phi(Y^t)\|_1\}}\nonumber\\
& +\sum_{t=1}^n\sum_{i=1}^d{\beta_i\|\phi(C^t(i))-\phi(Y^t)\|_1},
\end{align}


where $C^t(i)$ is the $i$-th colorized image of $f(X^t;\theta_f)$ and $\{\beta_i\}$ is a decreasing sequence. We use $d=4$ in our experiments.

The index of the best colorized video is not always the same. In most cases, we could empirically get a good index simply by choosing the one with the highest average per-pixel saturation where the saturation of a pixel is just the $S$ channel in the HSV color space. Our method could also be an interactive method for users to pick the results they want.

In Figure \ref{fig:diversity}, we show three colorized videos by our approach given the same grayscale input. In general, each video has its only style, and all the videos are different in both global color contrast and chrominance.

\section{Implementation}
We augment the input to the network $f$ by adding hypercolumn features extracted from the VGG-19 network \cite{simonyan2014very}. The hypercolumn features are expected to capture both low-level and high-level information of an image. In particular, we extract 'conv1\_2', 'conv2\_2', 'conv3\_2', 'conv4\_2' and 'conv5\_2' from the VGG-19 network and upsample the layers by bilinear upsampling to match the resolution of the input image. The total number of channels of the hypercolumn feature is 1472. 
We adopt U-Net \cite{Ronneberger2015} as our network structure for both networks $f$ and $g$, and modify the architecture to fit our purpose. We add a $1\times1$ convolutional layer at the beginning of each network to reduce the dimensionality of the input augmented with hypercolumn features \cite{Li2018}. To compute the optical flow, we use the state-of-the-art method PWC-Net \cite{Sun2018PWC-Net}.

For model training, we first train the network $f$ and then train $g$ and $f$ jointly. During each epoch for training $f$, we randomly sample 5,000 images in the ImageNet dataset \cite{Deng2009} to train with loss of $L_{bilateral}+L_{diversity}$ and sample 1,000 pairs of neighboring frames in the DAVIS training set \cite{Perazzi2016} by adding the temporal regularization for $f$, $L_{temporal}^f$. 
We train $f$ for 200 epochs in total. Then for training the refinement network $g$, we randomly sample 1,000 pairs of frames from the DAVIS dataset in each epoch with the loss $L_{temporal}^g$. While there are $d$ pairs of output from $f$ with diversity, we train $g$ on each pair of output. We also train our model in a coarse-to-fine fashion.  We train both networks on the 256p videos and images. Then we fine-tune our model on the 480p videos and images. 

\begin{table}[t!]
\centering
\setlength{\tabcolsep}{3mm}
\ra{1.25}
\begin{tabular}{@{}l@{\hspace{4mm}}c@{\hspace{3mm}}c@{}}
\toprule
& \multicolumn{2}{c}{Preference rate}\\
Comparison & DAVIS & Videvo \\
\midrule
Ours $>$ Zhang et al.\cite{Zhang2016} + BTC ~\cite{Lai2018} & 80.0\% &88.8\% \\
Ours $>$ Iizuka et al. \cite{Iizuka2016}+ BTC \cite{Lai2018} & 72.8\% & 63.3\% \\
\bottomrule
\end{tabular}
\vspace{2mm}
\caption{The results of perceptual user study. Both baselines are enhanced with temporal consistency by BTC \cite{Lai2018}. Our model consistently outperforms both state-of-the-art colorization methods by Zhang et al. \cite{Zhang2016} and Iizuka et al. \cite{Iizuka2016}. }
\label{table:user_study}
\end{table}

\section{Experiments}
\subsection{Experimental procedure}

\paragraph{Datasets.} 

We conduct our experiments mainly on the DAVIS dataset \cite{Perazzi2016} and the Videvo dataset \cite{videvo,Lai2018}. The test set of the DAVIS dataset consists of 30 video clips of various scenes. There are about 30 to 100 frames in each video clip. The test set of the Videvo dataset contains 20 videos and each one has about 300 video frames. In totally, we evaluate our models and baselines on 50 test videos. All the videos are resized to 480p in both datasets.

\paragraph{Baselines.} 
We compare our method with two state-of-the-art fully automatic image colorization approaches: the colorful image colorization (CIC) by Zhang et al. \cite{Zhang2016} and Iizuka et al.  \cite{Iizuka2016}. While these approaches are designed for image colorization, we apply their method to colorize video frame by frame. In addition, we apply the blind temporal consistency (BTC) method proposed by Lai et al. \cite{Lai2018} improve the overall temporal consistency. Lai et al. \cite{Lai2018} provided the results with temporal consistency for Zhang et al. \cite{Zhang2016} and Iizuka et al. \cite{Iizuka2016}. We use publicly available pre-trained models and results of the baselines for evaluation. Their pre-trained models are trained on the DAVIS dataset \cite{Perazzi2016} and the Videvo dataset \cite{videvo,Lai2018}.


\subsection{Results}
\paragraph{Perceptual experiments.}
To evaluate the realism of the colorized video by each method, we conduct a perceptual experiment by user study. We compare our method with Zhang et al.\cite{Zhang2016} and Iizuka et al. \cite{Iizuka2016} with enhanced temporal consistency by the blind temporal consistency (BTC) \cite{Lai2018}. While our approach generates multiple videos, we choose the video with high saturation for evaluation.

In the user study, there are video comparisons between our approach and a baseline. In each comparison, a user is presented with a pair of colorized 480p videos side by side. The user can play both videos multiple times. We set the order of video pairs randomly and let the user choose the one that is more realistic and temporally coherent. Totally 10 users participated in this user study.
\begin{figure*}
\centering
\begin{tabular}{@{}c@{\hspace{1mm}}c@{\hspace{1mm}}c@{\hspace{1mm}}c@{\hspace{1mm}}c@{}}
\rotatebox{90}{\small \hspace{2.5mm} With diversity}
&\includegraphics[width=0.24\linewidth]{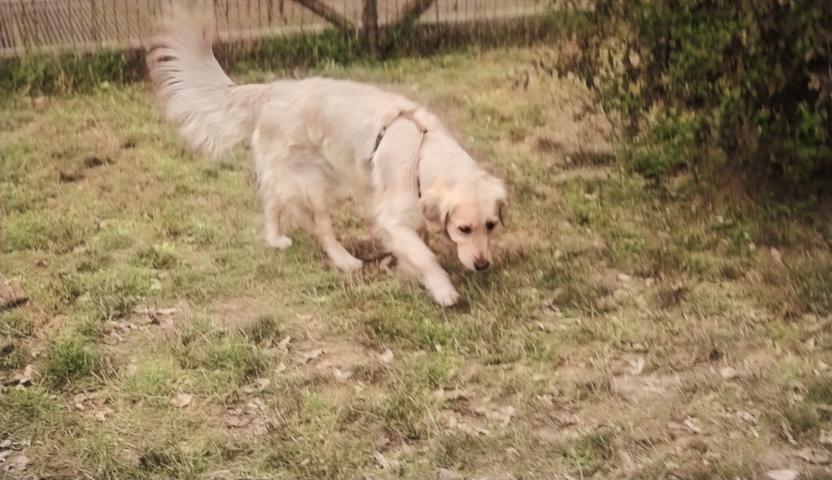}&
\includegraphics[width=0.24\linewidth]{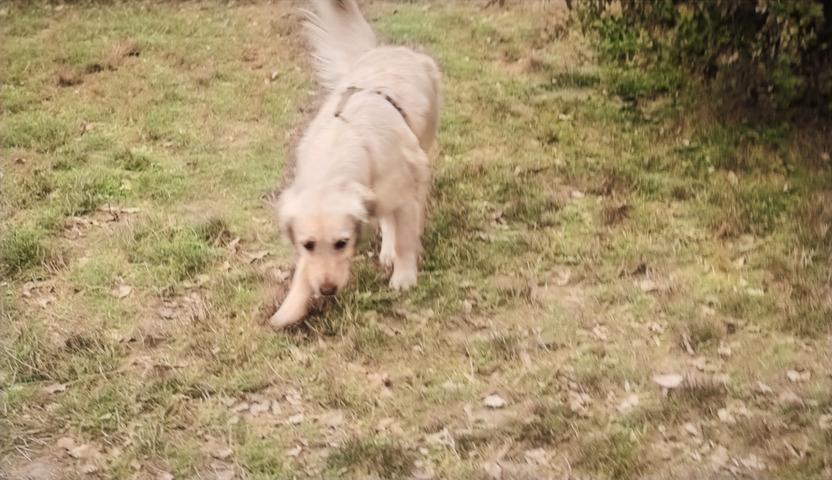}&
\includegraphics[width=0.24\linewidth]{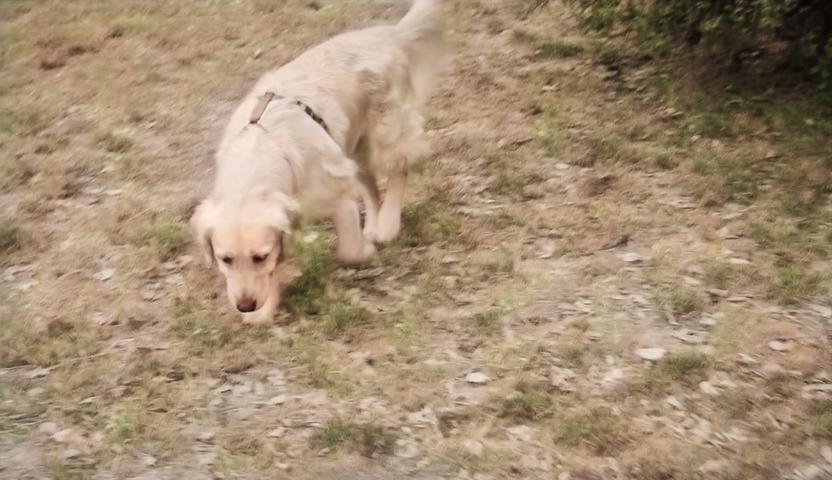}&
\includegraphics[width=0.24\linewidth]{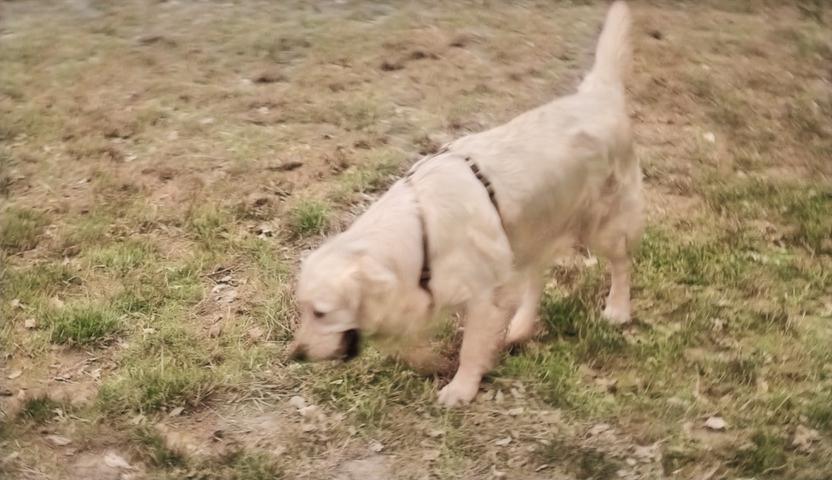}\\
\rotatebox{90}{\small \hspace{0.5mm} Without diversity}
&\includegraphics[width=0.24\linewidth]{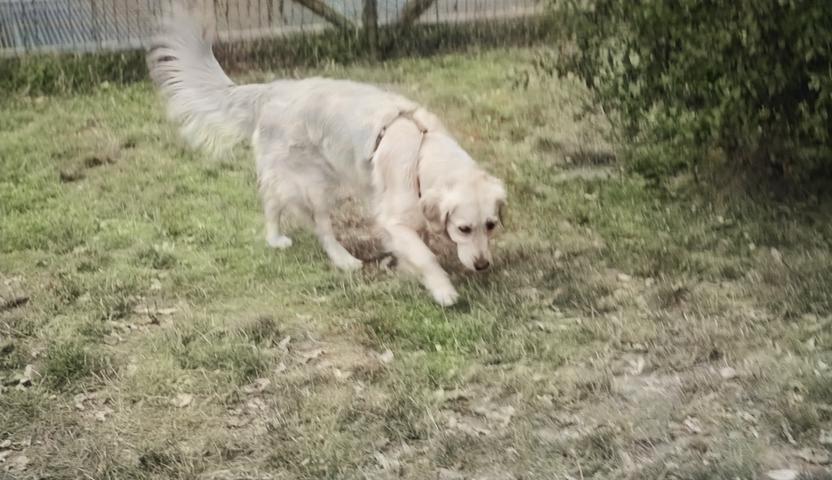}&
\includegraphics[width=0.24\linewidth]{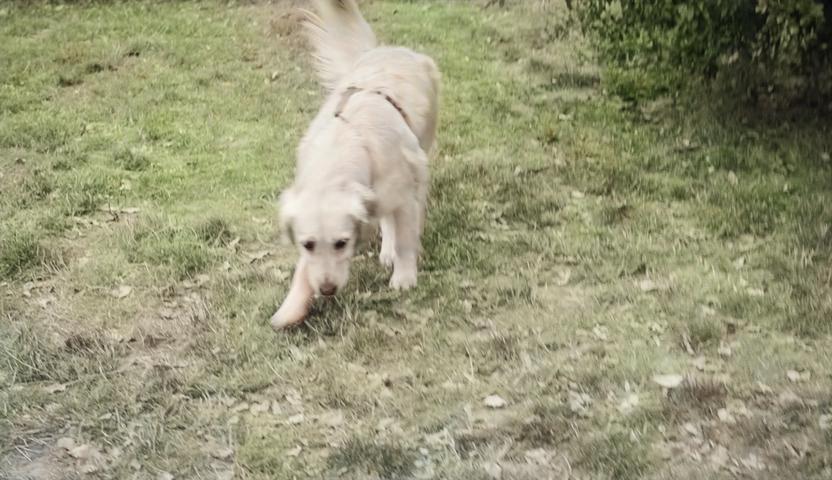}&
\includegraphics[width=0.24\linewidth]{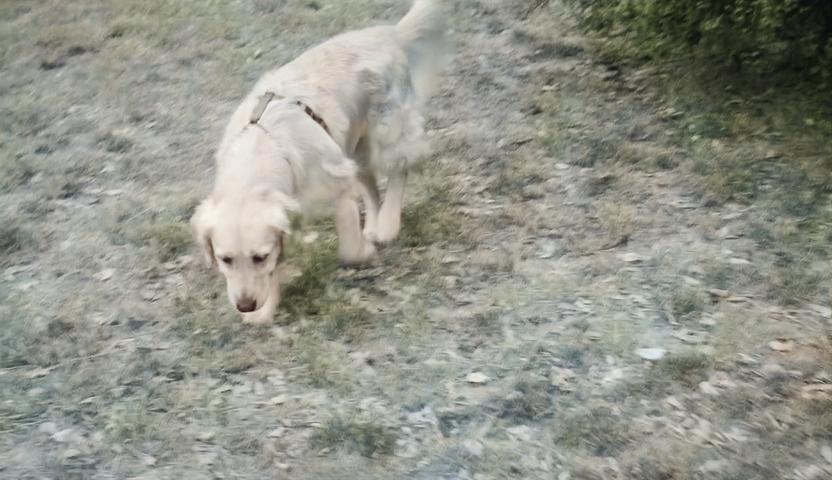}&
\includegraphics[width=0.24\linewidth]{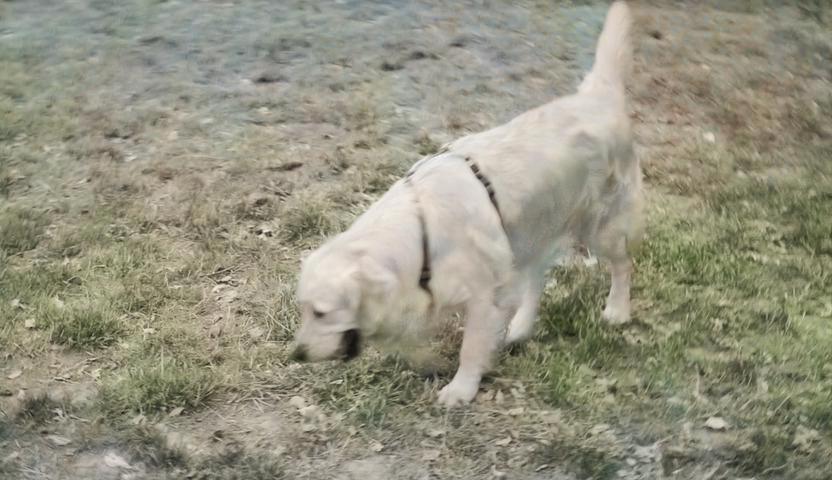}\\
&Frame 1 & Frame 2 & Frame 3 & Frame 4 \\
\end{tabular}
\vspace{1mm}
\caption{The visualization of the effect with and without the diversity loss. The first row shows four frames colorized by our full model, and the second shows four frames generated by our model without diversity. The diversity loss helps our model produce more temporally coherent and realistic results.}
\label{fig:Ablation-div}
\vspace{1mm}
\end{figure*}

\begin{figure}
\centering
\begin{tabular}{@{}c@{\hspace{1mm}}c@{\hspace{1mm}}c@{}}
\rotatebox{90}{\small \hspace{3mm} With self-reg.}
&\includegraphics[width=0.49\linewidth]{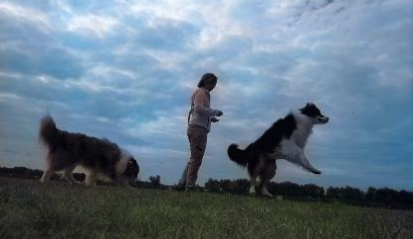}&
\includegraphics[width=0.49\linewidth]{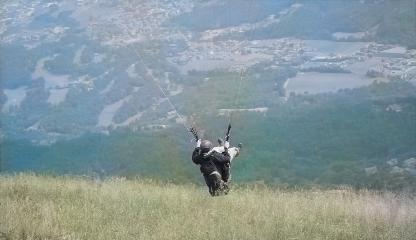}\\
\rotatebox{90}{\small \hspace{1mm} Without self-reg.}
&\includegraphics[width=0.49\linewidth]{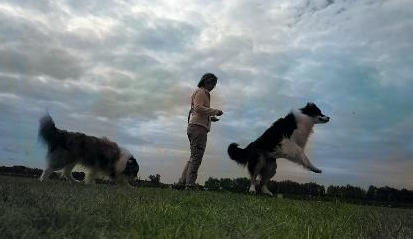}&
\includegraphics[width=0.49\linewidth]{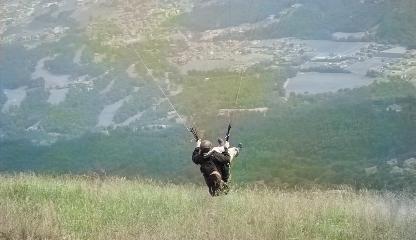}\\
&Video 1 & Video 2 \\
\end{tabular}
\vspace{1mm}
\caption{The visualization of the effect with and without the self-regularization. The self-regularization can help preserve global color consistency.}
\label{fig:Ablation-refine}
\vspace{1mm}
\end{figure}

Table \ref{table:user_study} summarizes the results of our perceptual experiment. Our method is consistently more rated preferable by most users. When our approach is compared with Zhang et al. \cite{Zhang2016}, our approach is preferred in 80.0\% of the comparisons on the DAVIS dataset and 88.8\% of the comparisons on the Videvo dataset \cite{videvo}. The perceptual user study is the key experiment to evaluate the performance of different methods.

\begin{table}[t!]
\centering
\setlength{\tabcolsep}{3mm}
\ra{1.25}
\begin{tabular}{@{}l@{\hspace{6mm}}c@{}}
\toprule
& Preference rate\\
Comparison & DAVIS \\
\midrule
Ours $>$ Ours without self-reg. & 67.9\%  \\
Ours $>$ Ours without diversity & 61.5\%  \\
\bottomrule
\end{tabular}
\vspace{3mm}
\caption{The results of the ablation study of comparisons between our full model and ablated models. The evaluation is performed by perceptual user study with 15 participants. The results indicate that self-regularization and diversity are key components in our model to achieve state-of-the-art performance in fully automatic video colorization.}
\label{table:ablation}
\end{table}
\paragraph{Ablation study.}
Table \ref{table:ablation} summarizes the ablation study by conducting perceptual user study on the DAVIS dataset. According to Table \ref{table:ablation}, our model without self-regularization or the diversity loss does not perform as well as our complete model. In summary, users rated our full model more realistic in 67.9\% of the comparisons between our full model and the model without self-regularization and in 61.5\% of the comparisons between our full model and the model without diversity.

\begin{table}
\centering
\setlength{\tabcolsep}{3mm}
\ra{1.25}
\begin{tabular}{@{}l@{\hspace{3mm}}c@{\hspace{3mm}}c@{\hspace{6mm}}c@{\hspace{3mm}}c@{}}
\toprule
& \multicolumn{2}{@{\hspace{-3mm}}c}{DAVIS} & \multicolumn{2}{c}{Videvo}\\
Method & LPIPS & PSNR & LILPS & PSNR \\
\midrule
Input & 0.227 & 23.80 & 0.228 & 25.30\\
Zhang et al.~\cite{Zhang2016} & 0.218 & 29.25 & 0.201 & 29.52\\
Iizuka et al.~\cite{Iizuka2016} & \textbf{0.189} & 29.91 & \textbf{0.190} & 30.23\\
Zhang et al. + BTC~\cite{Lai2018} & 0.243 & 29.07 & 0.249 & 29.04\\
Iizuka et al + BTC~\cite{Lai2018} & 0.218 & 29.25 & 0.241 & 28.90\\
Ours & 0.191 & \textbf{30.35} & 0.194 & \textbf{30.50}\\
\bottomrule
\end{tabular}
\vspace{2mm}
\caption{The results on two image similarity metrics, PSNR and LPIPS \cite{Zhang2018}. The blind temporal consistency (BTC) does not improve the results on these metrics. Image similarity metrics can not accurately measure the realism and temporal coherence of the colorized videos.}
\label{table:quant_other}
\end{table}

\paragraph{Qualitative results.} Figure \ref{fig:Ablation-div} and Figure \ref{fig:Ablation-refine} visualize the results of our full model and the ablated models without self-regularization or diversity. 

\begin{figure*}[p]
\centering
\begin{tabular}{@{}c@{\hspace{1mm}}c@{\hspace{1mm}}c@{\hspace{1mm}}c@{\hspace{1mm}}c@{\hspace{1mm}}c@{}}
\rotatebox{90}{\small \hspace{5mm} IZK}
&\includegraphics[width=0.19\linewidth]{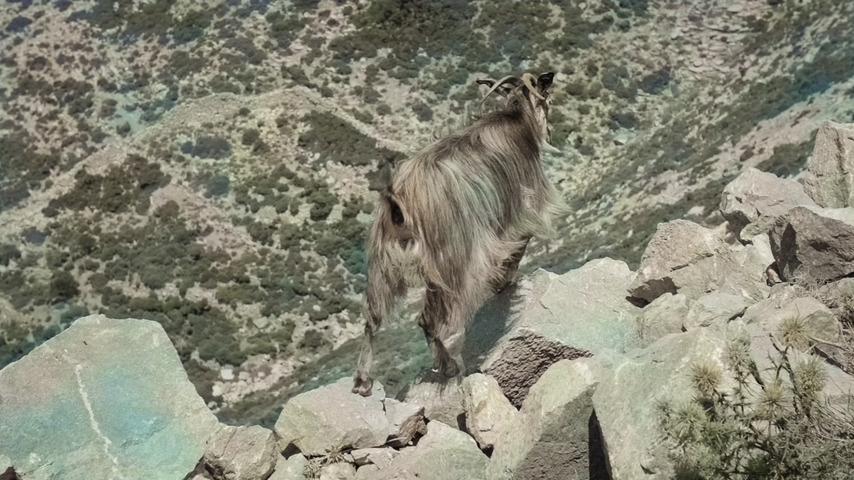}&
\includegraphics[width=0.19\linewidth]{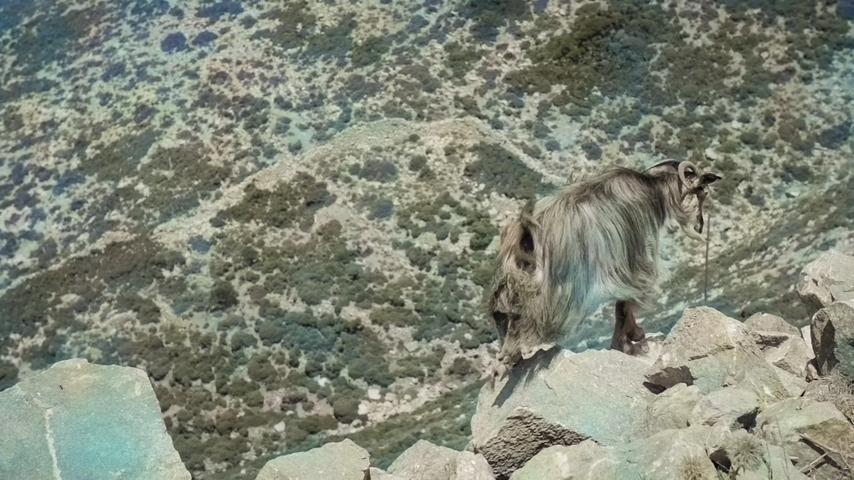}&
\includegraphics[width=0.19\linewidth]{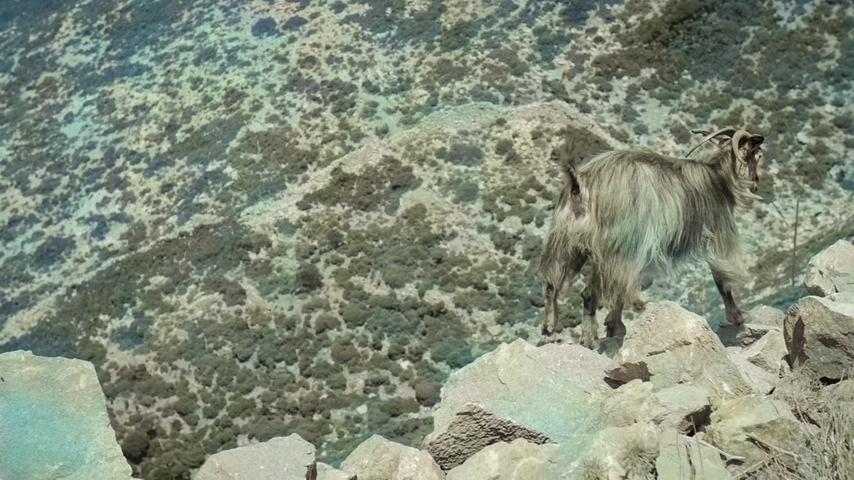}&
\includegraphics[width=0.19\linewidth]{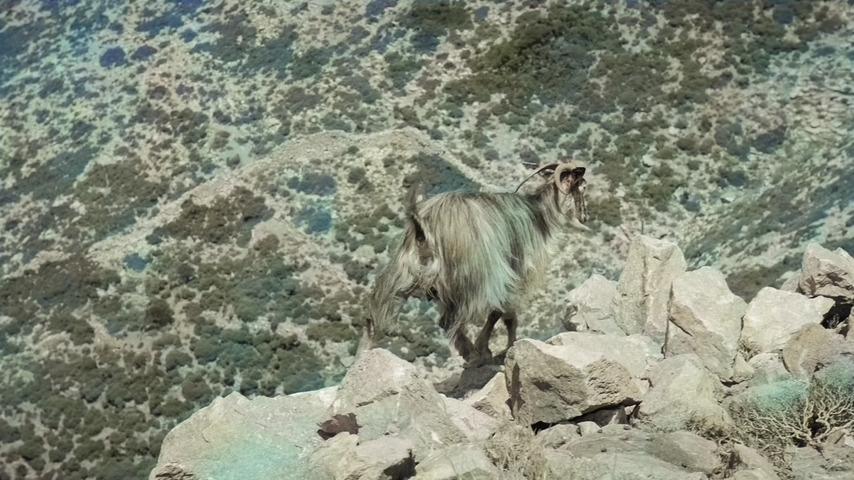}&
\includegraphics[width=0.19\linewidth]{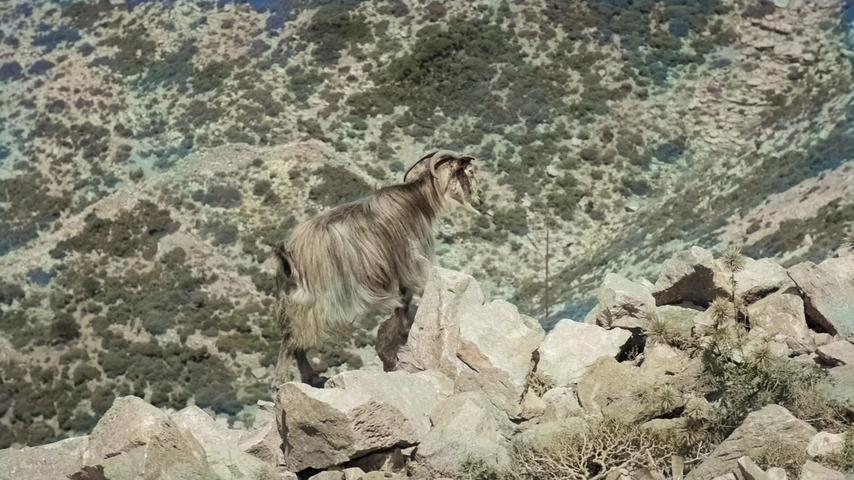}\\
\rotatebox{90}{\small \hspace{2mm}  IZK+BTC}
&\includegraphics[width=0.19\linewidth]{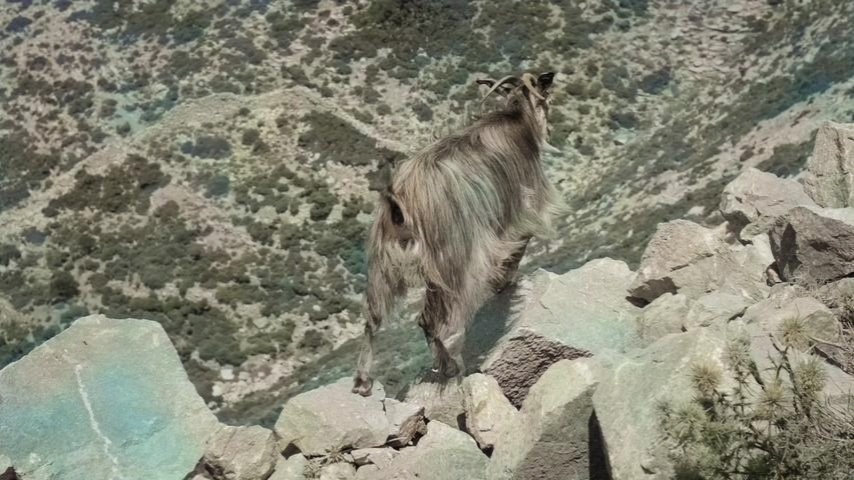}&
\includegraphics[width=0.19\linewidth]{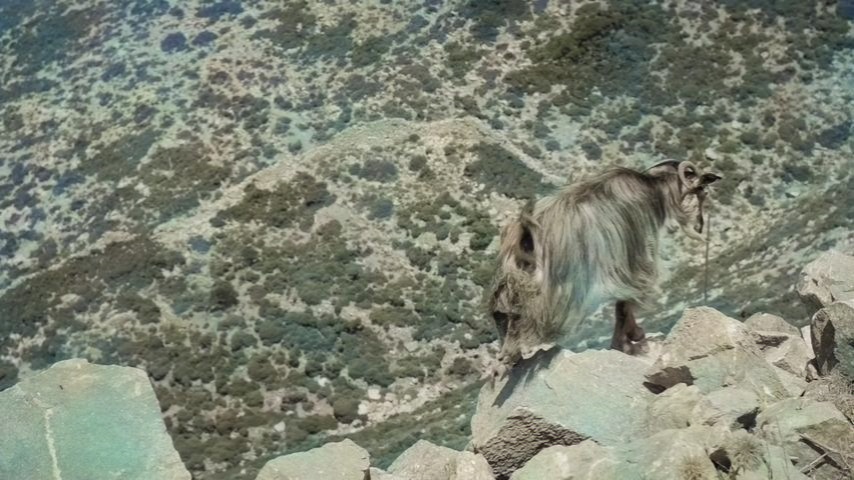}&
\includegraphics[width=0.19\linewidth]{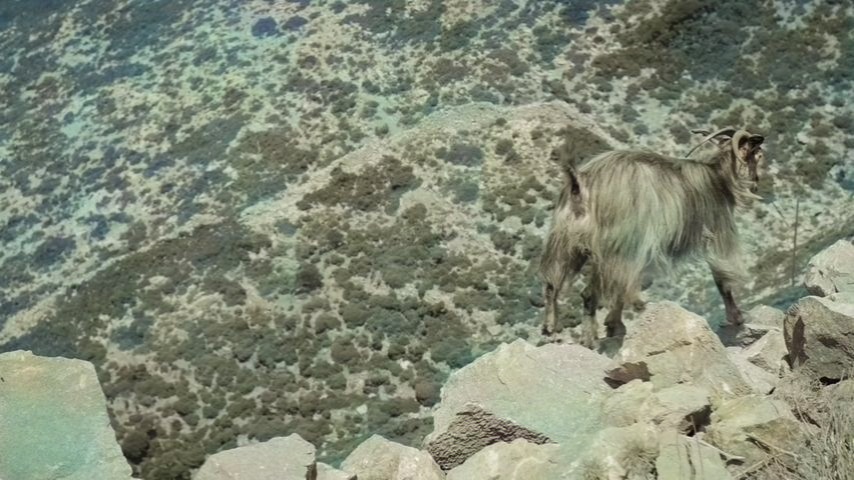}&
\includegraphics[width=0.19\linewidth]{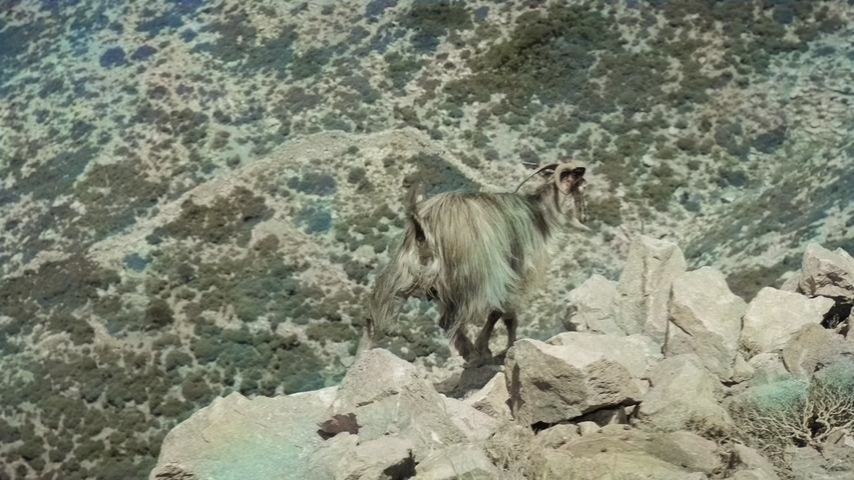}&
\includegraphics[width=0.19\linewidth]{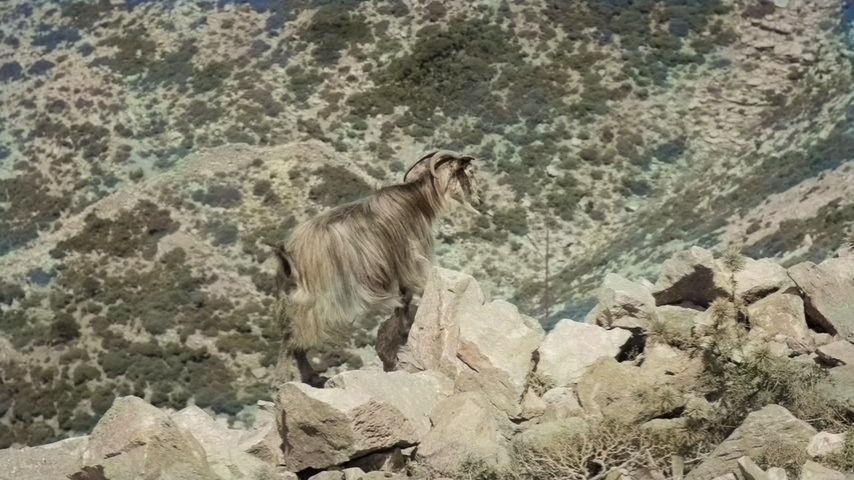}\\
\rotatebox{90}{\small \hspace{6mm} CIC}
&\includegraphics[width=0.19\linewidth]{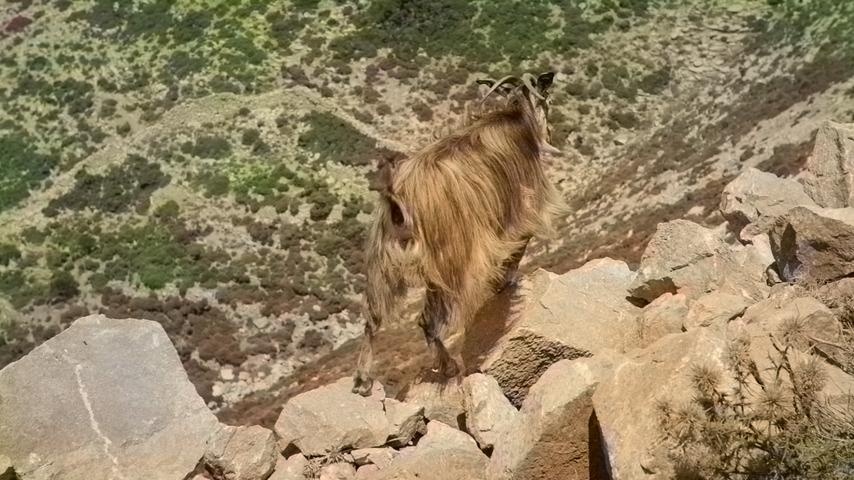}&
\includegraphics[width=0.19\linewidth]{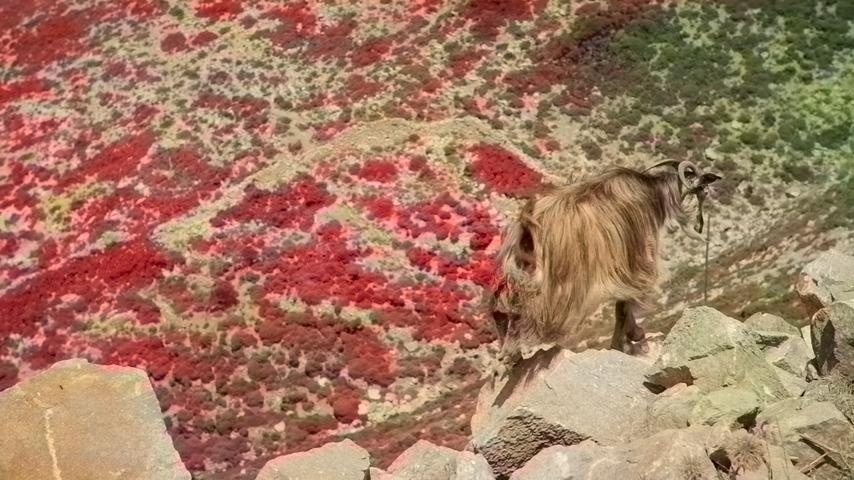}&
\includegraphics[width=0.19\linewidth]{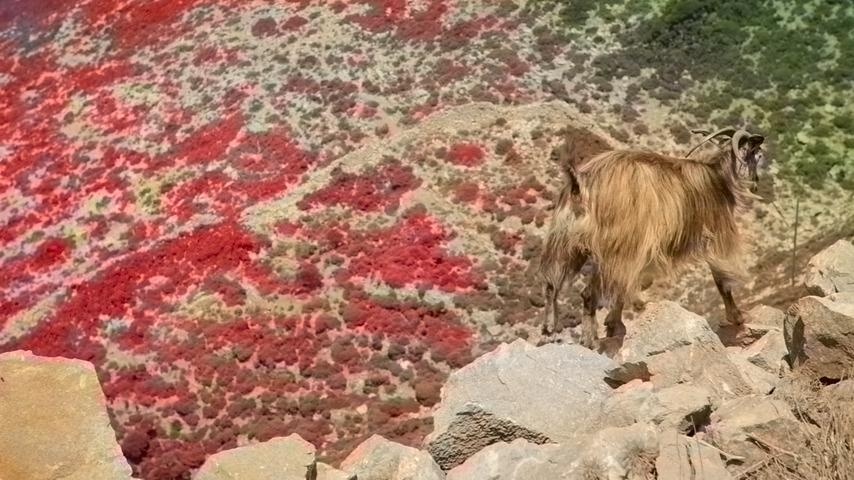}&
\includegraphics[width=0.19\linewidth]{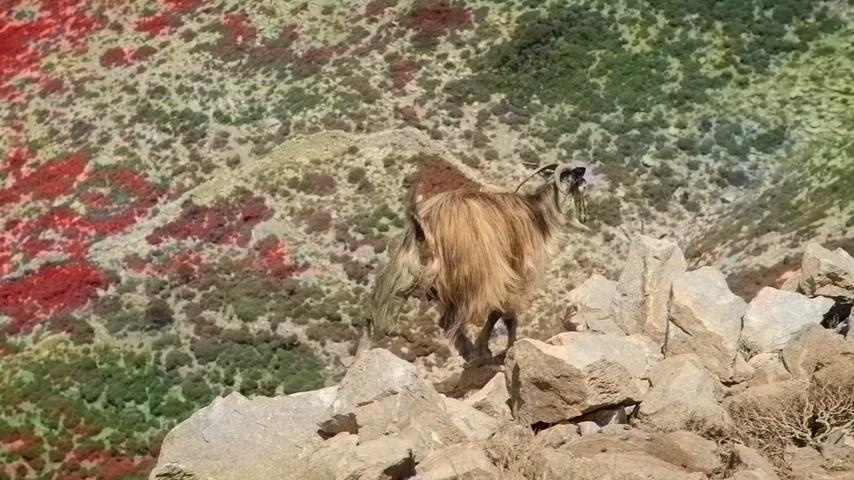}&
\includegraphics[width=0.19\linewidth]{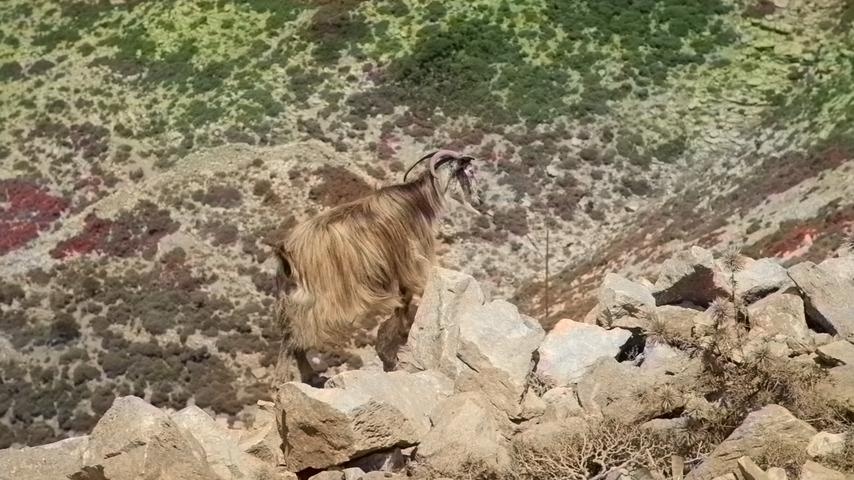}\\
\rotatebox{90}{\small \hspace{2mm}  CIC+BTC}
&\includegraphics[width=0.19\linewidth]{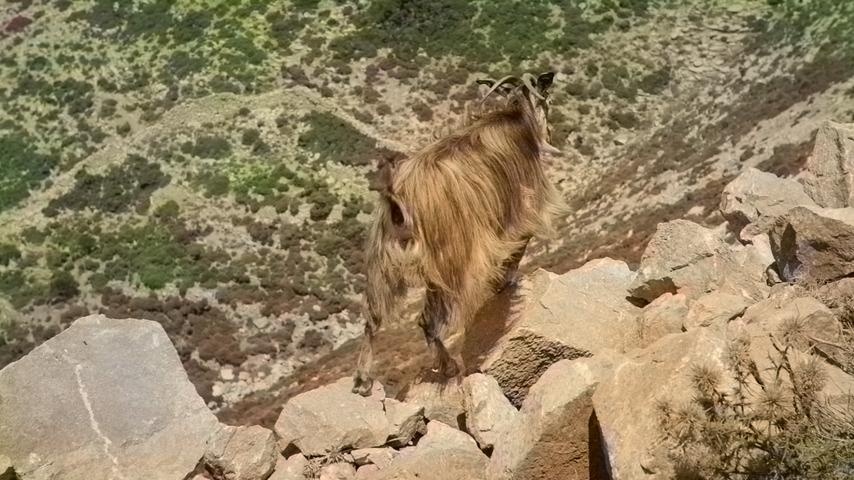}&
\includegraphics[width=0.19\linewidth]{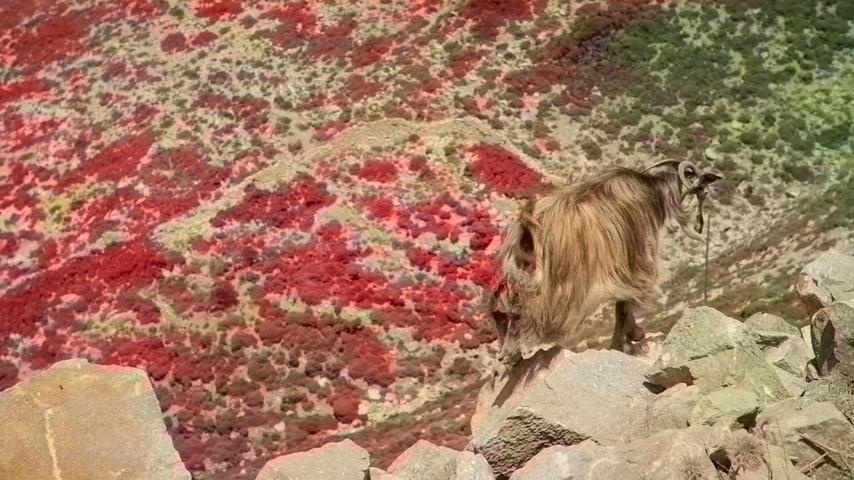}&
\includegraphics[width=0.19\linewidth]{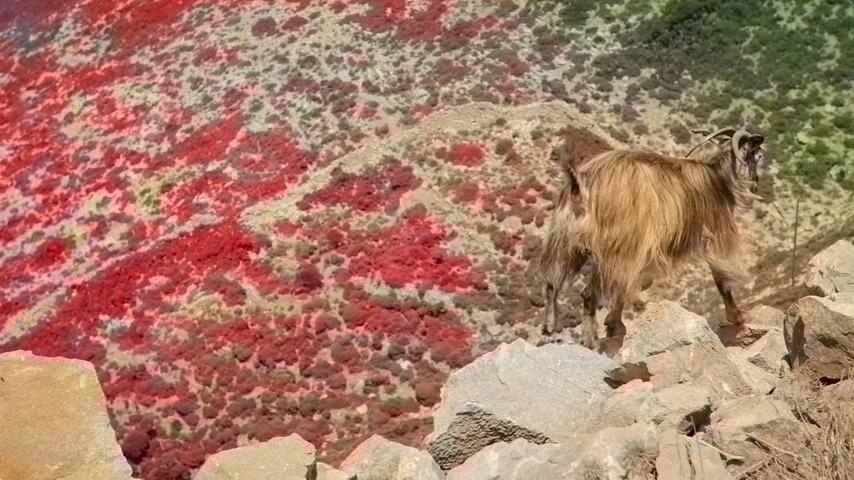}&
\includegraphics[width=0.19\linewidth]{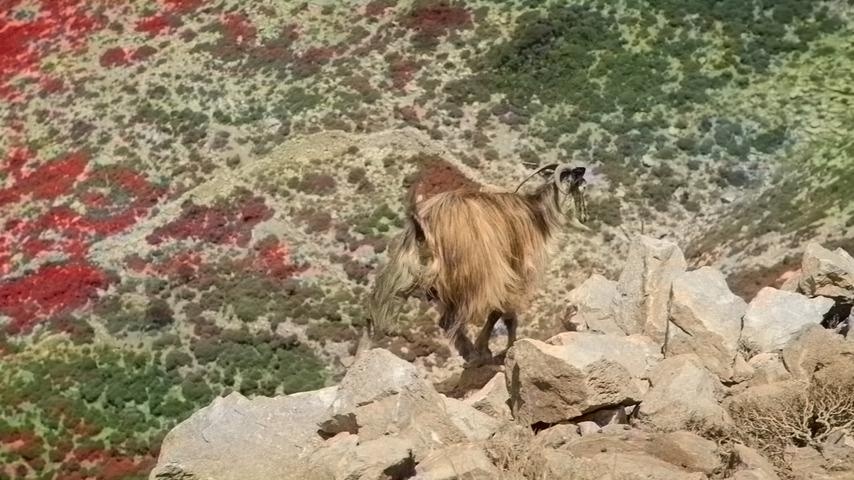}&
\includegraphics[width=0.19\linewidth]{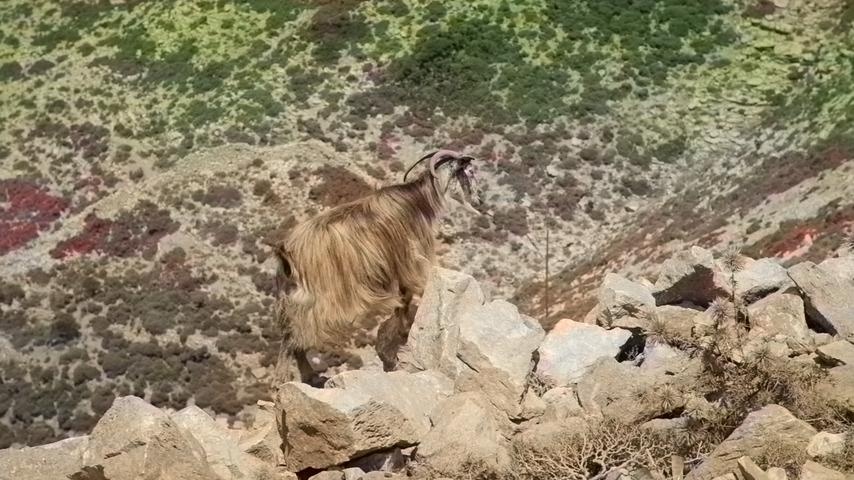}\\
\rotatebox{90}{\small \hspace{6mm} Ours}
&\includegraphics[width=0.19\linewidth]{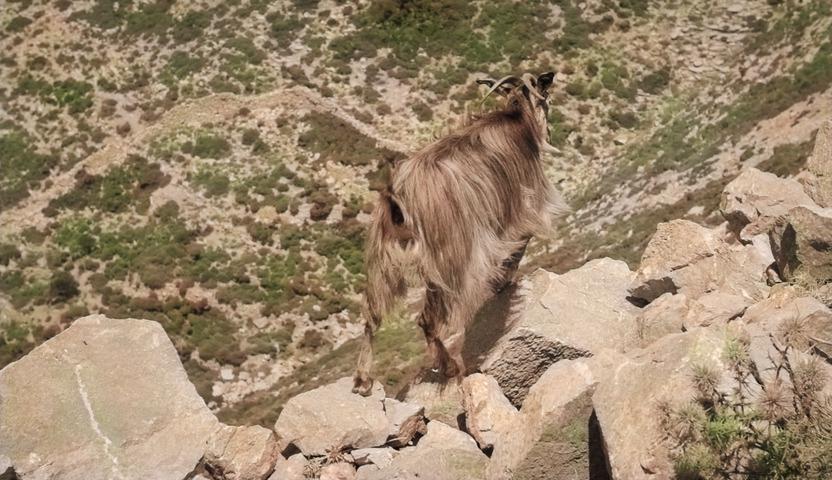}&
\includegraphics[width=0.19\linewidth]{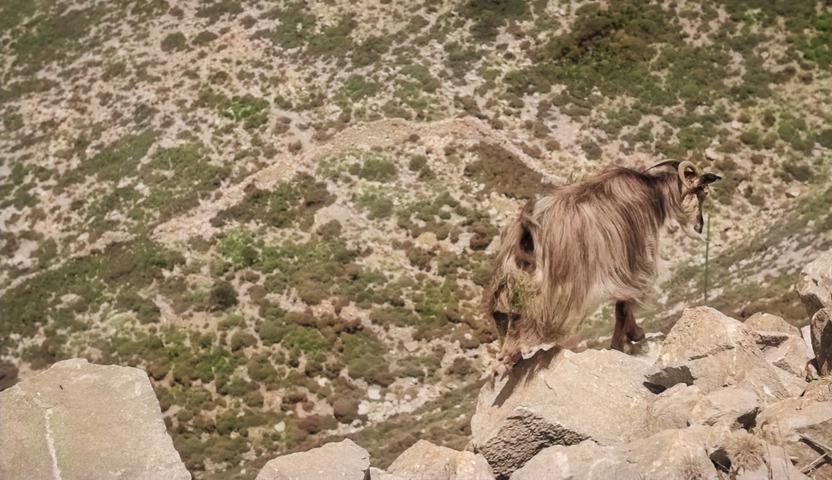}&
\includegraphics[width=0.19\linewidth]{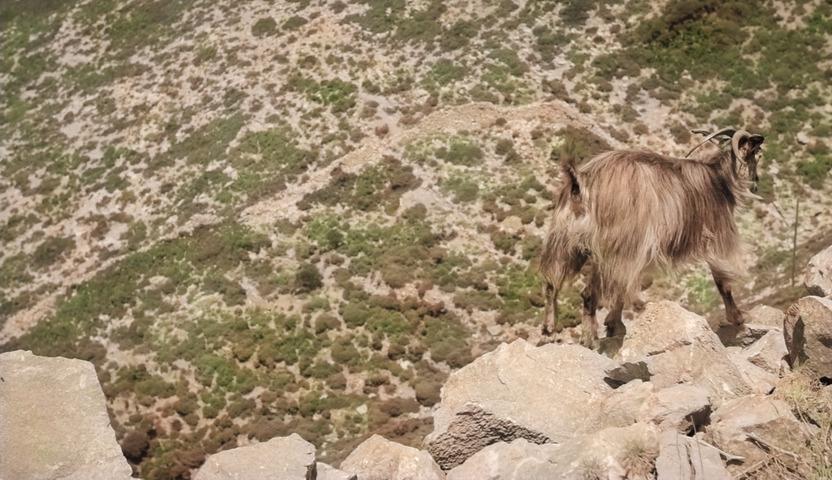}&
\includegraphics[width=0.19\linewidth]{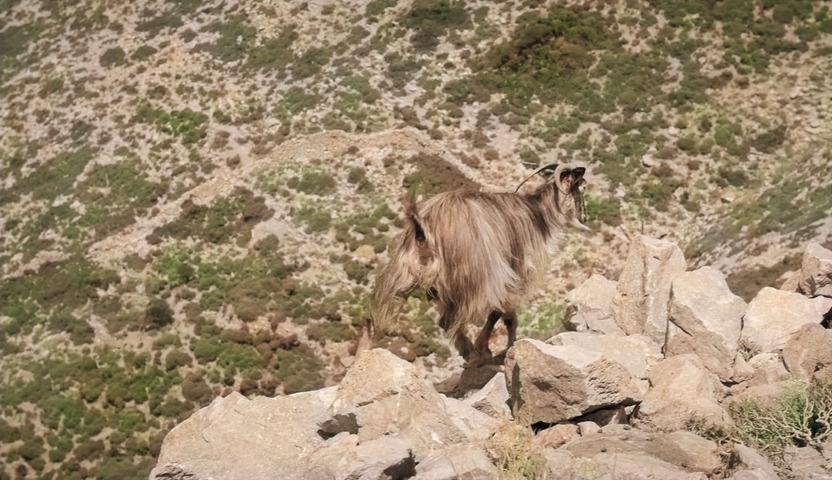}&
\includegraphics[width=0.19\linewidth]{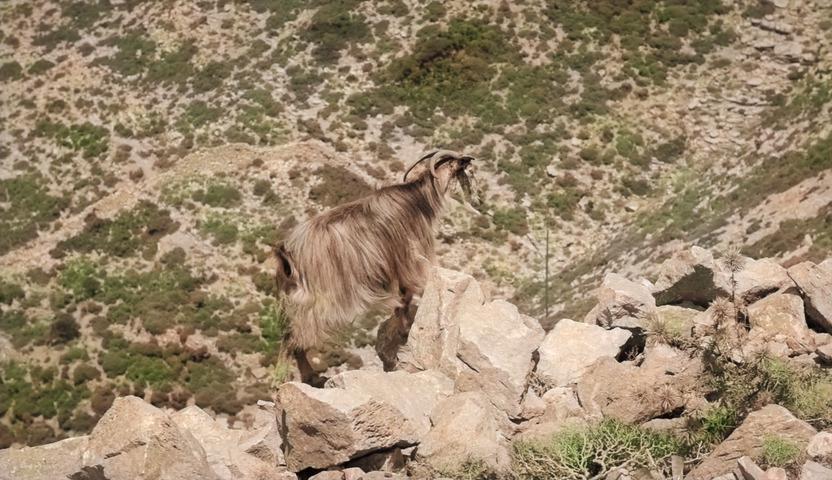}\\
&Frame 1 & Frame 2 & Frame 3 & Frame 4 & Frame 5 \\
\end{tabular}
\caption{Qualitative results on the DAVIS dataset \cite{Perazzi2016}. Here IZK refers to Iizuka et al. \cite{Iizuka2016}, CIC refers to the colorful image colorization method \cite{Zhang2016}, and BTC refers to the blind temporal consistency method \cite{Lai2018}. More results shown in the supplement.}
\label{fig:DAVIS}
\centering
\vspace{1mm}
\begin{tabular}{@{}c@{\hspace{1mm}}c@{\hspace{1mm}}c@{\hspace{1mm}}c@{\hspace{1mm}}c@{\hspace{1mm}}c@{}}
\rotatebox{90}{\small \hspace{5mm} IZK}
&\includegraphics[width=0.19\linewidth]{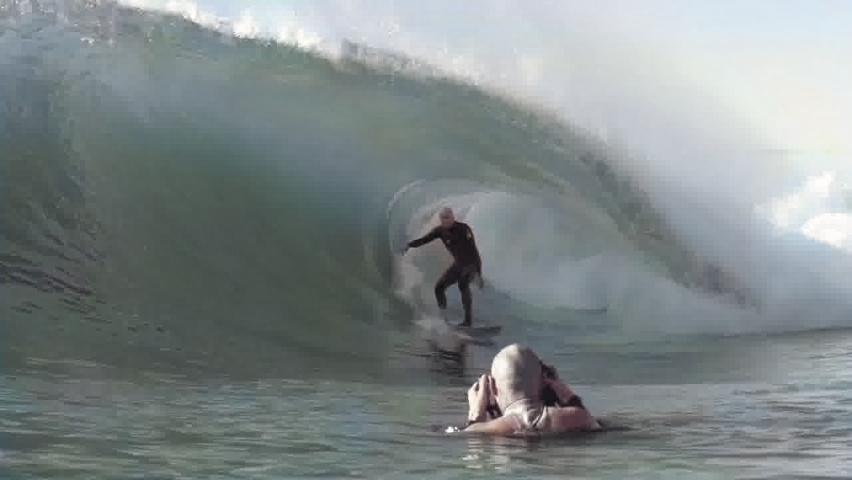}&
\includegraphics[width=0.19\linewidth]{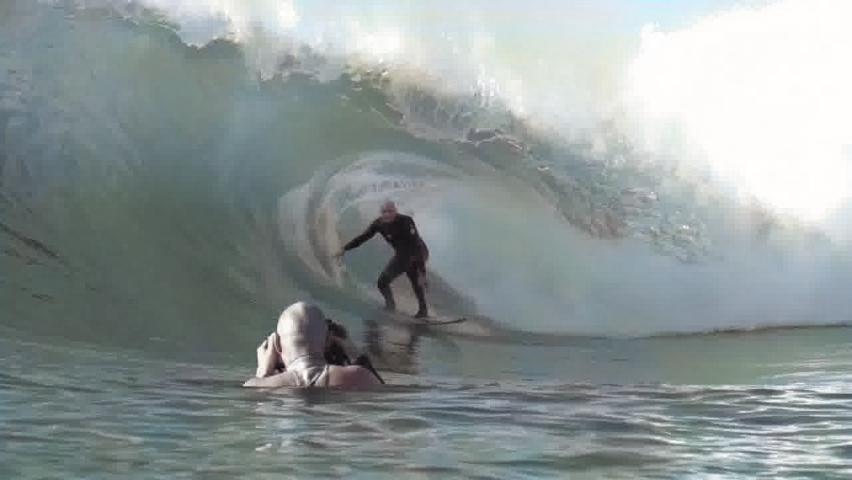}&
\includegraphics[width=0.19\linewidth]{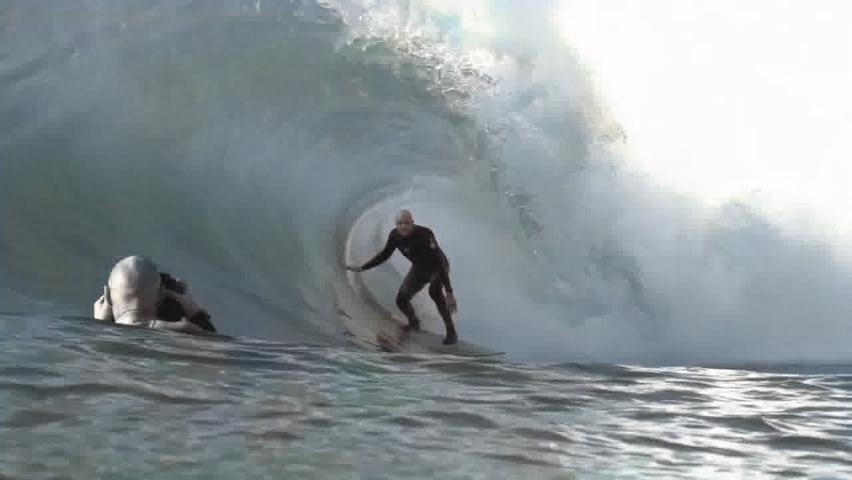}&
\includegraphics[width=0.19\linewidth]{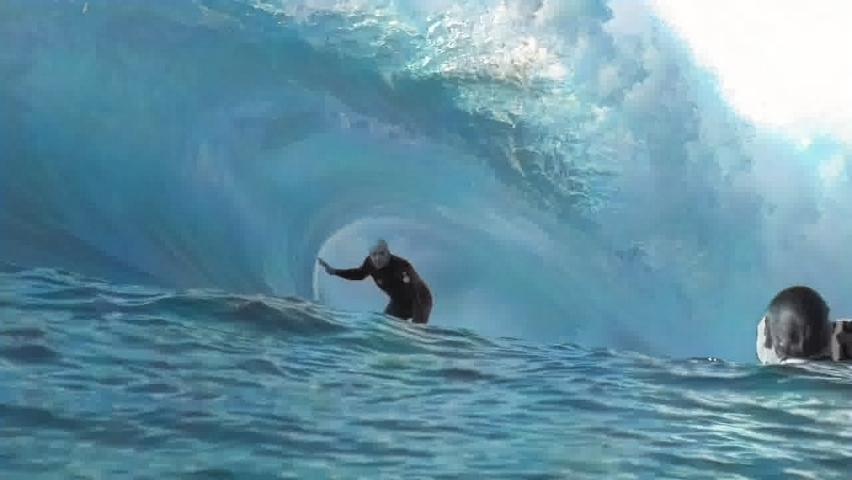}&
\includegraphics[width=0.19\linewidth]{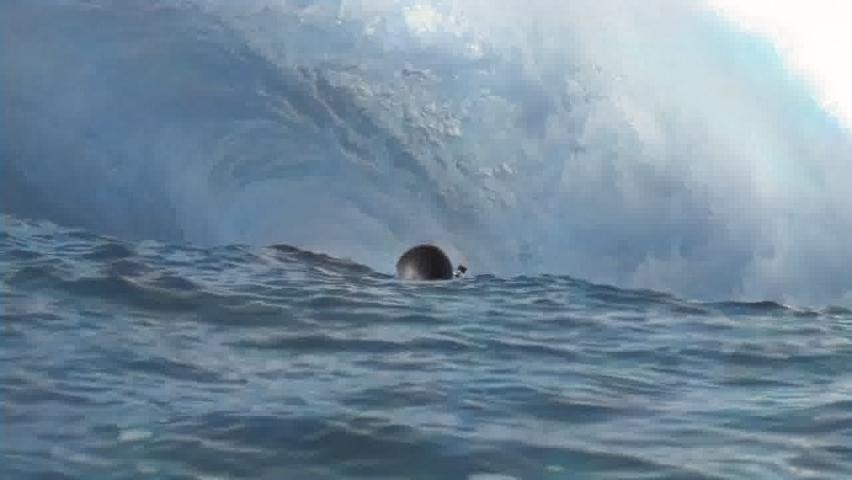}\\
\rotatebox{90}{\small \hspace{2mm} IZK+BTC}
&\includegraphics[width=0.19\linewidth]{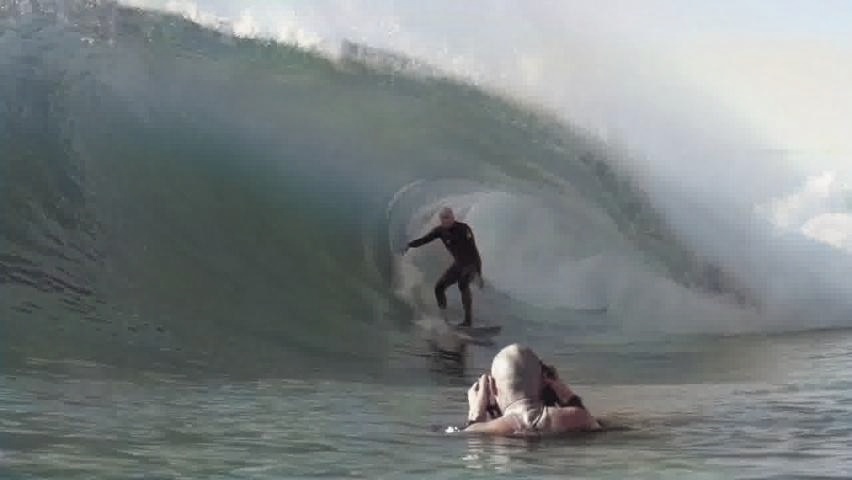}&
\includegraphics[width=0.19\linewidth]{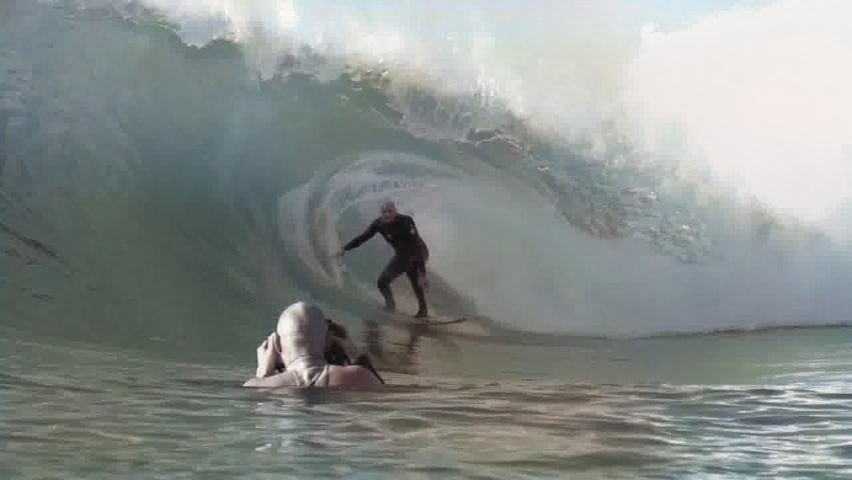}&
\includegraphics[width=0.19\linewidth]{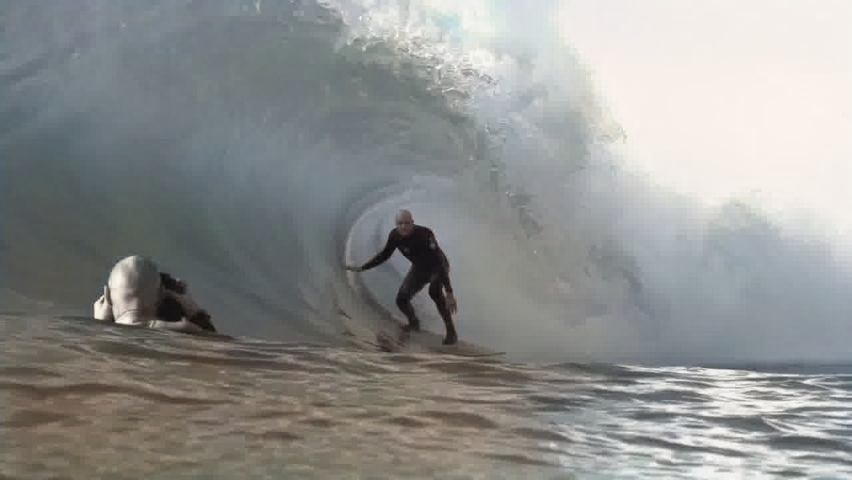}&
\includegraphics[width=0.19\linewidth]{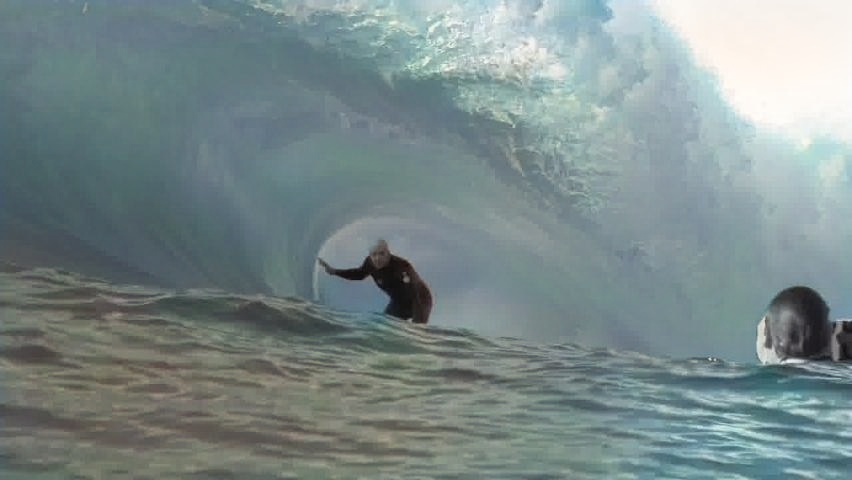}&
\includegraphics[width=0.19\linewidth]{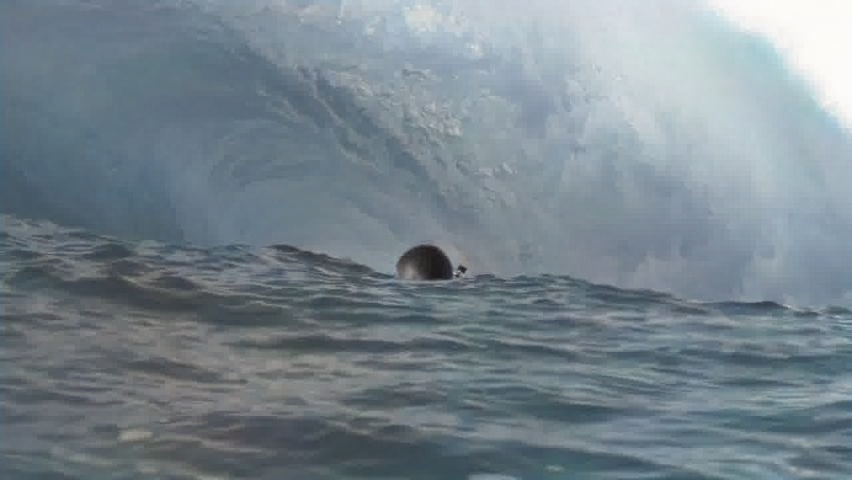}\\
\rotatebox{90}{\small \hspace{6mm} CIC}
&\includegraphics[width=0.19\linewidth]{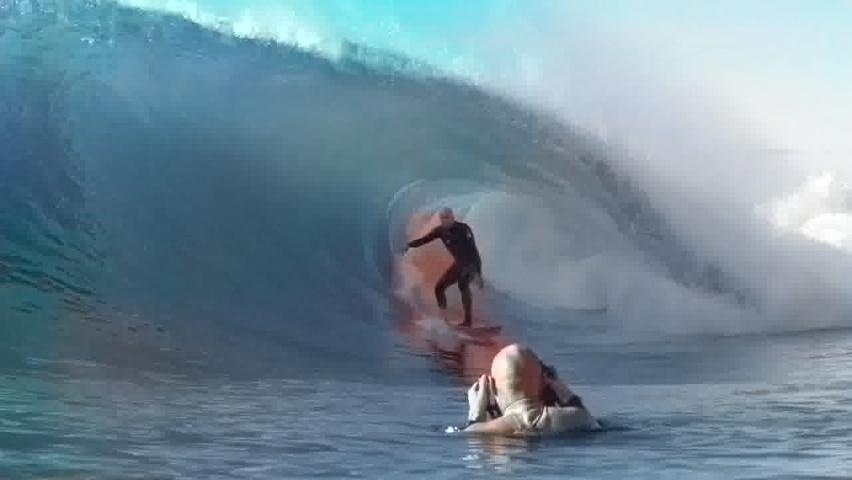}&
\includegraphics[width=0.19\linewidth]{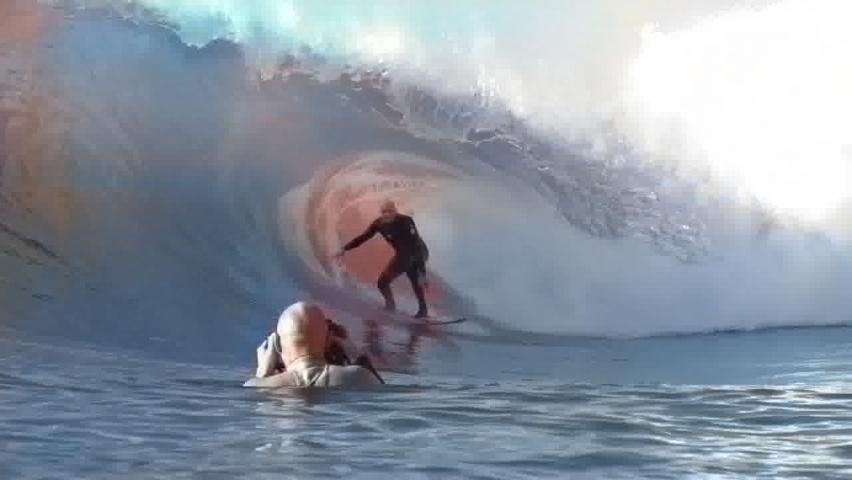}&
\includegraphics[width=0.19\linewidth]{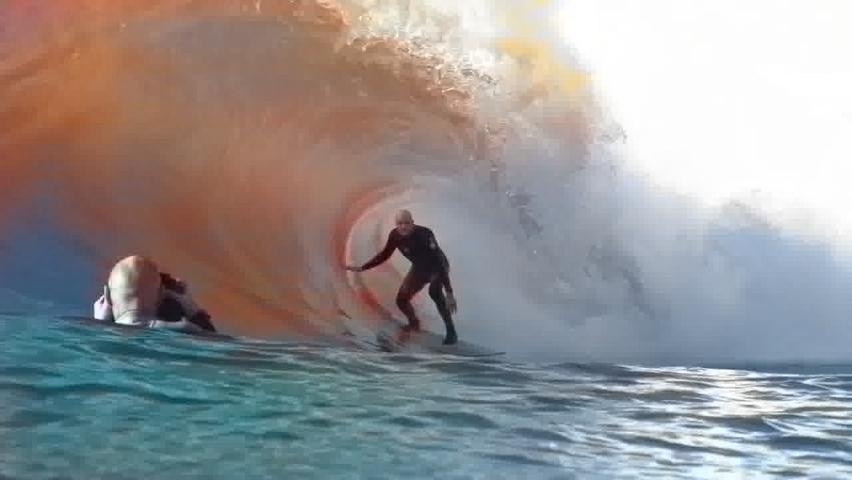}&
\includegraphics[width=0.19\linewidth]{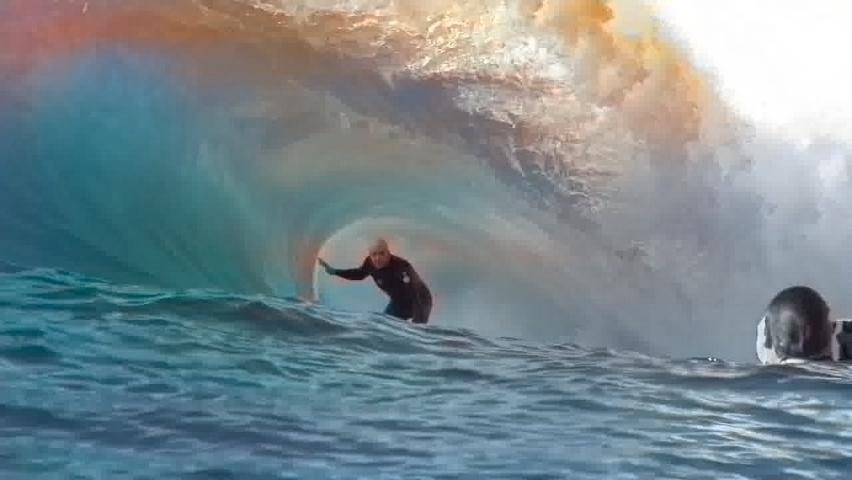}&
\includegraphics[width=0.19\linewidth]{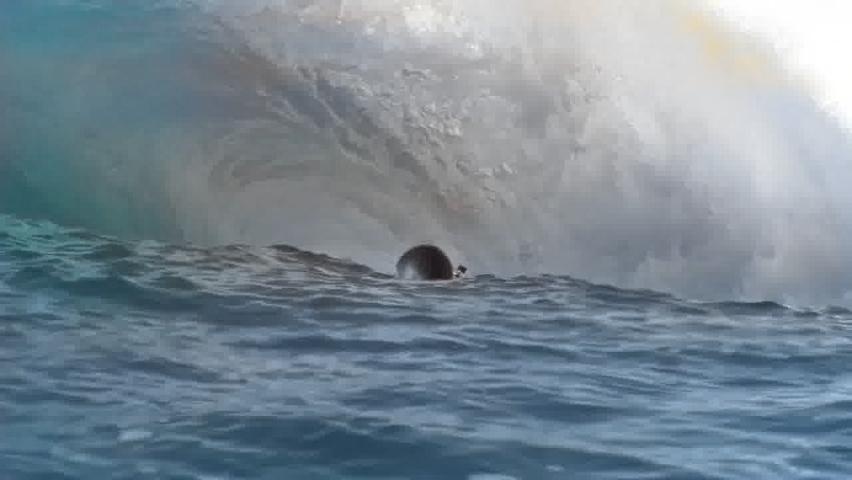}\\
\rotatebox{90}{\small \hspace{2mm} CIC+BTC}
&\includegraphics[width=0.19\linewidth]{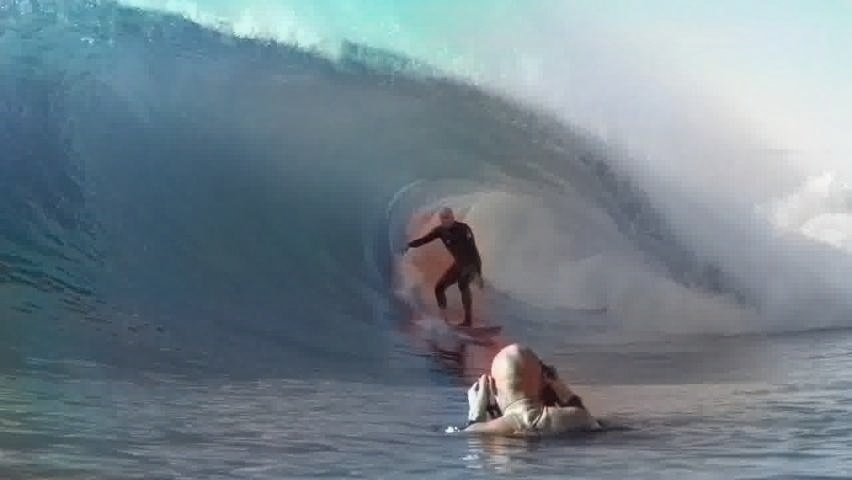}&
\includegraphics[width=0.19\linewidth]{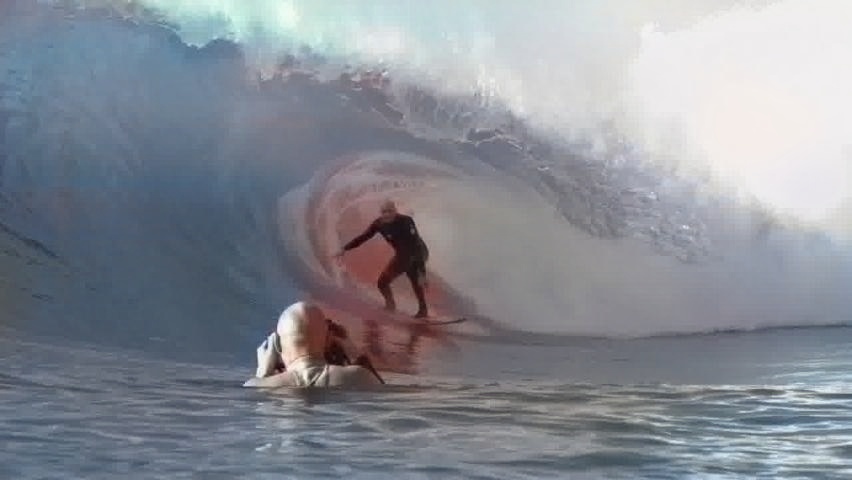}&
\includegraphics[width=0.19\linewidth]{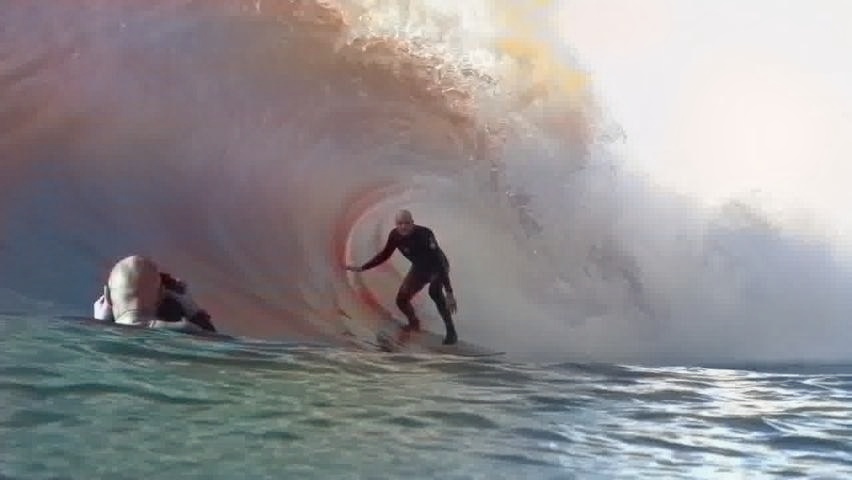}&
\includegraphics[width=0.19\linewidth]{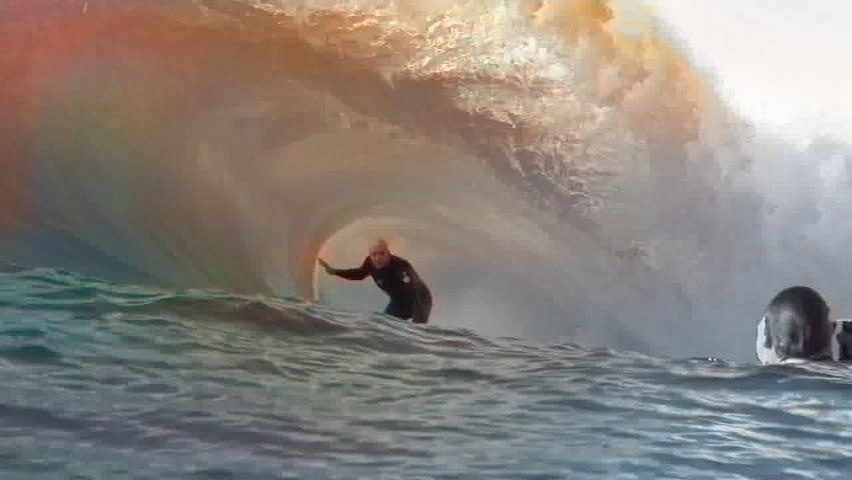}&
\includegraphics[width=0.19\linewidth]{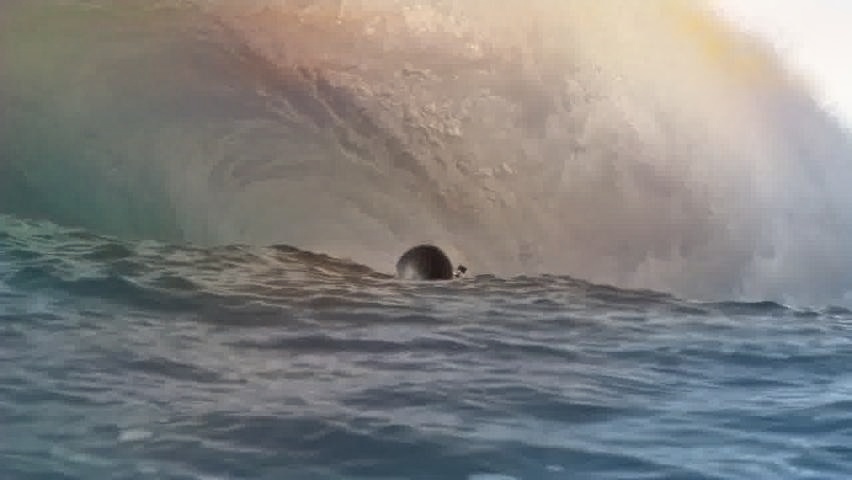}\\
\rotatebox{90}{\small \hspace{6mm} Ours}
&\includegraphics[width=0.19\linewidth]{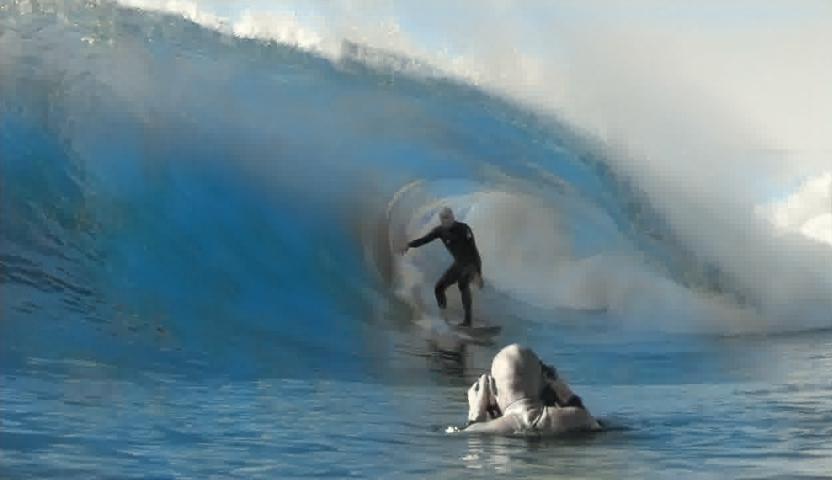}&
\includegraphics[width=0.19\linewidth]{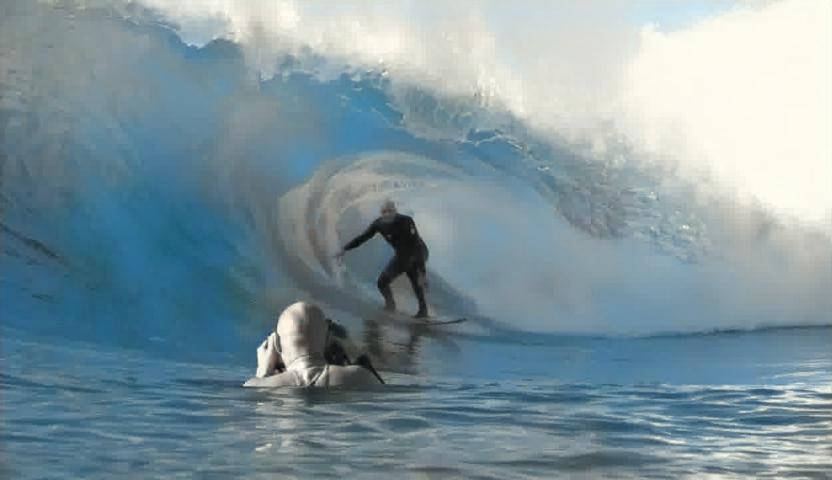}&
\includegraphics[width=0.19\linewidth]{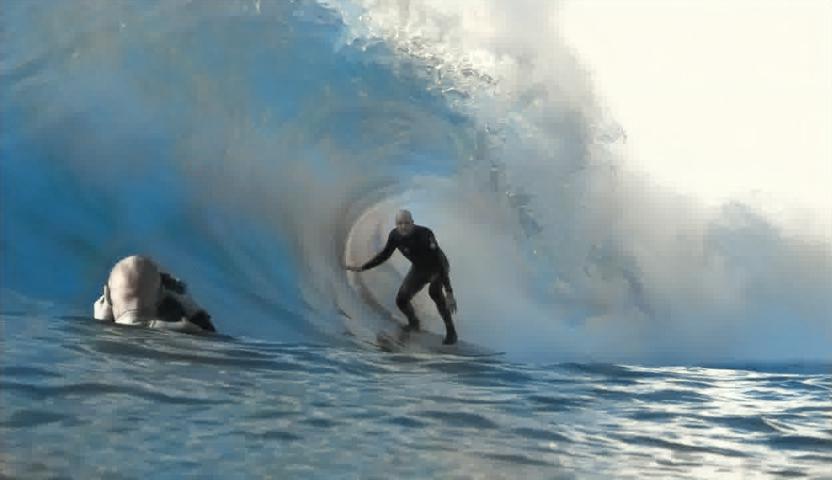}&
\includegraphics[width=0.19\linewidth]{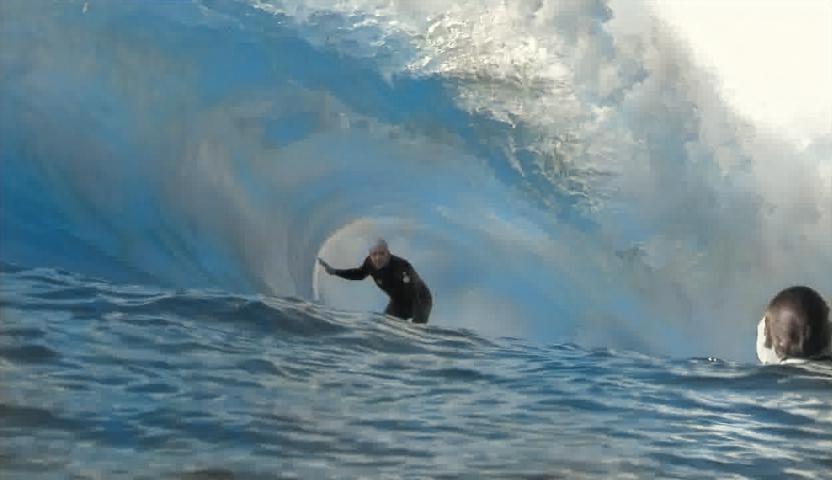}&
\includegraphics[width=0.19\linewidth]{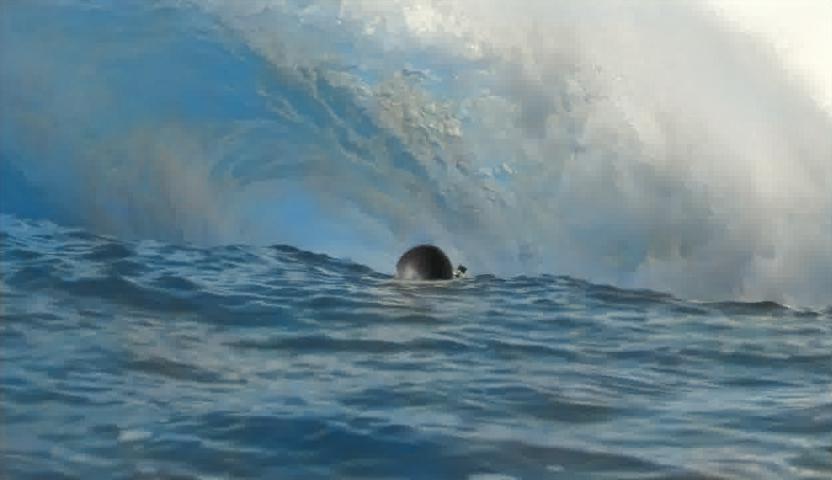}\\
&Frame 1 & Frame 2 & Frame 3 & Frame 4 & Frame 5 \\
\end{tabular}
\caption{Qualitative results on the Videvo dataset \cite{videvo}. Here IZK refers to Iizuka et al. \cite{Iizuka2016}, CIC refers to the colorful image colorization method \cite{Zhang2016}, and BTC refers to the blind temporal consistency method \cite{Lai2018}. More results shown in the supplement.}
\label{fig:Videvo}
\end{figure*}

In Figure \ref{fig:DAVIS} and Figure \ref{fig:Videvo}, we show the result videos  colorized by our method and prior work. Our method produces more temporally consistent and more realistic colorized videos than state-of-the-art approaches do.

\paragraph{Image similarity metrics.} We can use the image similarity metrics as a proxy to measure the similarity between the colorized video and the ground-truth video. Table \ref{table:quant_other} summarizes the results on image similarity metrics. Note that these metrics do not directly reflect the degree of realism of colorized videos. For example, a car may be colorized as blue or red. Both colors are plausible choices, but choosing a color different from the ground-truth video can results in huge errors on these image similirity metrics. 


\section{Discussion}
We have presented our fully automatic video colorization model with self-regularization and diversity. Our colorized videos preserve global color consistency in both bilateral space and temporal space. By utilizing a diversity loss, our model is able to generate a diverse set of colorized videos that differentiate different modes in the solution space. We also find that our diversity loss stabilizes the training and process. Our work is an attempt to improve fully automatic video colorization but the results are still far from perfect. We hope our ideas of self-regularization and diversity can inspire more future work in fully automatic video colorization and other video processing tasks.

{\small
\bibliographystyle{ieee}
\bibliography{egpaper_for_review}
}
\end{document}